\documentclass[aps,prb,reprint,amsmath,amssymb,floatfix,superscriptaddress,longbibliography]{revtex4-2}
\usepackage[T1]{fontenc}
\usepackage[utf8]{inputenc}
\usepackage[bookmarks=false,linkcolor=NavyBlue,urlcolor=NavyBlue,colorlinks,citecolor=NavyBlue]{hyperref}
\usepackage{graphicx}
\usepackage{amsmath}
\usepackage[svgnames]{xcolor}
\usepackage{xcolor}
\usepackage{enumerate}
\usepackage[normalem]{ulem}
\usepackage{booktabs}
\usepackage{orcidlink}
\usepackage{comment}
\usepackage{longtable}

\graphicspath{{./Figures/}}

\newcommand{\figref}[2]{\hyperref[#1]{\ref{#1}(#2)}}
\newcommand{\figsref}[2]{\hyperref[#1]{\ref{#1}#2}}

\newcommand{\E}{\mathbb{E}}

\newcommand{\Var}{\mathbb{V}\mathrm{ar}}

\newcommand{\eps}{\epsilon}
\newcommand{\bs}[1]{\boldsymbol{#1}}

\begin{document}

\title{Neural networks become centroid classifiers when trained on heavily corrupted inputs}

\newcommand{\AEPCornell}{Department of Applied and Engineering Physics, Cornell University, Ithaca, NY 14853, USA}
\newcommand{\PhysicsCornell}{Department of Physics, Cornell University, Ithaca, NY 14853, USA}

\author{Justin Tahmassebpur}
\thanks{These authors contributed equally to this work.}
\affiliation{\AEPCornell}
\author{Asadullah Bhuiyan}
\thanks{These authors contributed equally to this work.}
\affiliation{\PhysicsCornell}
\author{Hyejin Kim}
\affiliation{\PhysicsCornell}
\author{Omri Lesser}
\affiliation{\PhysicsCornell}


\begin{abstract}
Learning from imperfect data is a central theme in machine learning, connecting practical questions of robustness to fundamental questions of learnability. Here we examine attribute noise: learning from corrupted inputs while keeping the labels intact, a setting that has received considerably less analytical attention than its label-noise counterpart. We consider two types of corruption models: additive noise and replacement noise. Through experiments with multi-layer perceptrons (MLPs) on corrupted classification datasets, we find that neural networks remain robust, maintaining well-above-chance accuracy even when inputs are $>90\%$ corrupted---far beyond human recognition. To understand this robustness, we analyze infinite-width networks in the heavy-corruption regime using a mean-field-inspired approach and derive a leading-order decision rule for the classification outcome: the network implements a prototype rule, the nearest-class-mean, assigning each test point to the class whose training-set average it most closely resembles. This leading-order decision rule is universal across a broad range of MLP architectures, holding for any depth, as well as a wide class of activation functions and noise distributions. The same centroid mechanism closely matches finite-width network behavior in our experiments and provides an interpretable and analytically tractable account of why learning can succeed even when individual training examples carry almost no signal.
\end{abstract}

\maketitle
\section{\label{sec:intro}Introduction}

Machine learning is fundamentally reshaping science and technology, raising profound questions about both its capabilities and its limitations.
A central challenge in this context is learning from imperfect data: how can a model generalize when the examples available during training are degraded?
The canonical setting is that of \emph{label noise}, where the target labels in a classification or regression task are corrupted~\cite{angluin1988noisy,kearns1998statistical,blum2003noise,frenay2014survey,natarajan2013noisy,rolnick2017robust,northcutt2021confident}.
The complementary problem of \emph{attribute noise}---where the labels remain correct but the input features themselves are corrupted---has received considerably less analytical attention. Furthermore, attribute noise is common in practice: measured attributes are often degraded by sensor noise, acquisition errors, missing features, or other imperfections, while the associated labels may remain reliable~\cite{emmanuel2021survey}.

\begin{figure*}
    \centering
    \includegraphics[width=\linewidth]{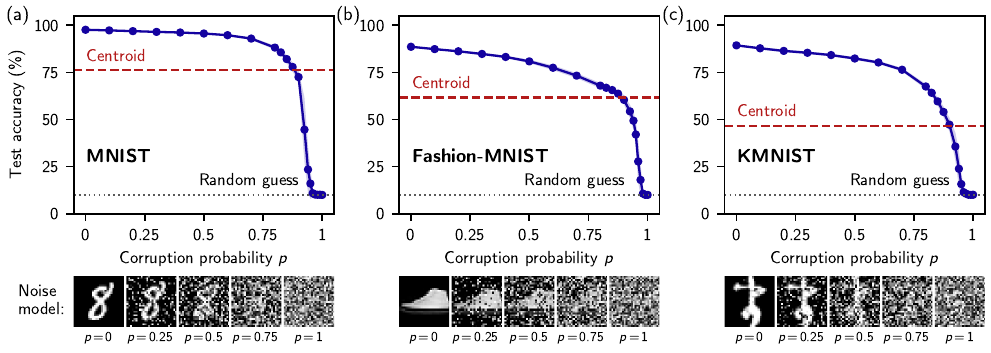}
    \caption{Test accuracy vs.\ corruption probability $p$ for width-128, 3-layer ReLU MLPs trained with cross-entropy loss on images with i.i.d.\ pixel replacement noise (blue solid lines). Red dashed lines show dataset-averaged test accuracy computed from the centroid classifier Eq.~\eqref{eq:maintxt_classification_rule} on the clean data---we provide this as a baseline. Three class-balanced datasets are shown: (a)~MNIST, (b)~Fashion-MNIST, and (c)~KMNIST. Below each panel we demonstrate the noise model by showing representative examples of images subject to replacement noise at increasing $p$. For each $p$, an MLP is trained on a dataset where each pixel is independently replaced with uniform noise with probability $p$; the test set is left clean. 
    Accuracy metrics are averaged over 100 independent noisy training realizations per $p$ value, and shaded bands surrounding each curve represent one standard deviation across realizations (see Appendix~\ref{app:test_acc_var} for a discussion of the non-monotonic behavior of the standard deviation).
    Even when the replacement noise has corrupted up to 80\% ($p=0.8$) of each image in the training set, test accuracy remains at or above ${\sim}70\%$ across all three datasets, with accuracy collapsing toward uniform random guessing ($100/C = 10\%$ for $C=10$ classes, dashed line) only near $p \to 1$. Furthermore, the centroid baseline agrees with the noisy test accuracy near $p\sim0.9$.
    }
    \label{fig:MNIST+}
\end{figure*}

Attribute noise has been examined from several perspectives.
Early work showed that injecting noise into inputs during training can act as a regularizer and improve generalization~\cite{matsuoka1992noise,holmstrom1992additive,bishop1995noise,an1996effects,sietsma1991creating,reed1992jittered,zur2009noise,ramirez2025role}.
Later work addressed feature corruption and deletion more directly, developing classifiers that are robust to missing inputs and training procedures that explicitly marginalize over corrupted features~\cite{globerson2006nightmare,dekel2009missing,vandermaten2013marginalized}.
A separate line of research considers \emph{test-time} corruption, asking how models trained on clean data degrade when evaluated on corrupted inputs~\cite{hendrycks2020augmix}.
This form of robustness is quantified by benchmarks such as ImageNet-C~\cite{hendrycks2019benchmarking} and MNIST-C~\cite{mu2019mnistc}.
Together, these bodies of work address several important aspects of the robustness of corruption.
However, they leave open a basic question: can neural networks trained on \emph{severely} corrupted attributes, but clean labels, learn a coherent classification rule and generalize well to clean test data?

In this paper, we show that the answer is surprisingly ``yes''.
Empirically, we train multi-layer perceptrons (MLPs) on corrupted versions of the standard image datasets MNIST~\cite{lecun2010mnist}, KMNIST~\cite{clanuwat2018deep}, and Fashion-MNIST~\cite{xiao2017fashion}, using two corruption models appropriate for image data: pixel replacement noise and additive noise.
Across all three datasets, we find that the networks remain strikingly robust even when the training inputs are severely corrupted; see Fig.~\ref{fig:MNIST+}.
At such high-noise levels, the images are nearly indistinguishable to the human eye, yet the network can still learn to classify clean test images far better than chance.

We then explain this robustness analytically using the neural tangent kernel (NTK) framework~\cite{jacot2018ntk}.
For any wide MLP trained on arbitrary datasets with clean labels and generic independent and identically distributed (i.i.d.) input corruption, we derive an explicit analytical form for the learned predictor in the high-noise limit.
The resulting high-noise decision rule is simple: the analytical form of the predictor is a linear nearest-class-mean---or \textit{centroid}~\cite{tibshirani2002diagnosis}---classifier on top of a competing stochastic background.

This theoretical result is then compared to numerical experiments on finite-width MLPs, and we find excellent agreement for both raw logit outputs and the $\mathrm{argmax}$-filtered test accuracy.
This empirical agreement with finite-width networks suggests that their surprising robustness on the image datasets considered here comes not from a complicated denoising or feature-recovery procedure, but from aggregating the weak class information that survives the corruption into a prototype-based decision rule---the centroid rule.

Furthermore, the leading form of the high-noise decision rule is independent of depth, activation, and the details of the chosen noise distribution.
The origin of this universality can be understood through the lens of symmetry and linear response: at pure noise, training examples become exchangeable, and the leading class-dependent response to a small surviving non-exchangeable signal is forced to depend on centered class averages.

These results connect our work to prototype-based views of classification, which have appeared in both classical and modern settings, including nearest-class-mean methods, prototypical networks, and recent analyses of class-mean structure in deep representations~\cite{tibshirani2002diagnosis,snell2017prototypical,guerriero2018deepncm,papyan2020neuralcollapse,seddik2022classwise,kothapalli2023neuralcollapse}.
Our setting differs in emphasis: we do not assume a prototype-based model from the outset, nor do we study prototype geometry as an empirical tendency of learned representations.
Instead, we \emph{derive} the prototype rule analytically as the effective predictor of a wide network trained in the high-attribute-noise regime.
More broadly, our results identify a rare regime in which neural networks can be understood in unusually explicit terms: despite the complexity of the architecture, the learned rule itself becomes simple, universal, and interpretable.

The remainder of this paper is organized as follows.
In Sec.~\ref{sec:numerics}, we introduce the two corruption models and present our motivating empirical finding: MLPs classify clean test data well above chance even with heavy train-time corruption.
In addition, we present results on MLP classification with independent test-time \textit{and} train-time corruption, finding that classification power is optimized when the training set corruption strength matches that of the test set.
Following this, we derive a theoretical prediction for the trained network output in the high-noise regime using the NTK framework and compare with numerical experiments on finite-width MLPs in Sec.~\ref{sec:analytics}.
Furthermore, we present a symmetry argument for the origin of the decision rule and its universality across noisy MLP architectures.
Finally, Sec.~\ref{sec:discussion} discusses the implications of our results and outlook.
Details on numerical experiments, technical derivations, and supporting details are collected in the Appendices.

\section{Feature corruption models and empirical results}\label{sec:numerics}
In this section, we present an empirical study of multiclass classification with MLPs trained on feature-corrupted inputs. We will first present the corruption channels of interest, followed by a presentation of empirical results for MLP performance when trained on correctly-labeled but feature-corrupted inputs, and tested on clean data; this setup will be treated analytically in Sec.~\ref{sec:analytics}. In addition, we briefly present empirical results for MLP performance when both the training \textit{and} testing examples are independently corrupted, with further details provided in Appendix~\ref{app:ptrain_ptest}.

Empirical findings presented in this section were obtained from fully-connected neural networks with 3 hidden layers of width 128, rectified linear unit (ReLU) activation, trained for 20 epochs using the Adam optimizer with batch size 128 and cross-entropy loss, and averaged over 100 random seeds. Further details on the experiments can be found in Appendix~\ref{app:experiment-details}.

\subsection{Corruption models and setup}
We begin by introducing the notation and noise models used throughout this work.
Let \(x_{k\chi}\) denote the \(k\)-th feature of \emph{clean} training example \(\chi\), where \(k=1,\ldots,d\) indexes the feature dimension and \(\chi=1,\ldots,N\) indexes the training set.
The corresponding corrupted data point is denoted by \(\tilde{x}_{k\chi}\). Our main theoretical analysis focuses on the large-feature-dimension regime \(d\gg 1\), which is natural for image data and many modern high-dimensional learning settings.

\begin{figure}
    \centering
    \includegraphics[width=\linewidth]{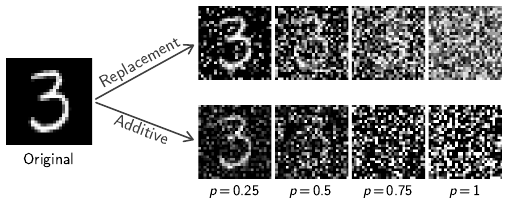}
    \caption{Visualization of the two corruption models used in this work applied to an MNIST image. The top panel shows the replacement noise model, where each pixel is independently replaced with uniform noise with probability $p$; see Eq.~\eqref{eq:replacement_noise}. The bottom panel shows the additive noise model, where each pixel is replaced by a convex combination of the clean pixel and an independent noise variable, with mixing parameter $p$; see Eq.~\eqref{eq:additive_noise}. The panels show increasing corruption levels $p$.}
    \label{fig:noise_models_viz}
\end{figure}

We consider two corruption models, as shown in Fig.~\ref{fig:noise_models_viz}. The first is \emph{replacement noise},
\begin{equation}\label{eq:replacement_noise}
    \tilde{x}_{k\chi} = (1-b_{k\chi})x_{k\chi} + b_{k\chi}u_{k\chi},
\end{equation}
where \(b_{k\chi}\) are i.i.d.\ Bernoulli random variables with
\(\Pr(b_{k\chi}=1)=p\), so that each feature (each pixel in image data) is independently replaced with probability \(p\), and \(u_{k\chi}\) is an i.i.d.\ noise variable.
In our numerical experiments we take \(u_{k\chi}\sim \mathrm{Unif}[0,1]\), which is a natural choice for grayscale image data such as MNIST, where each of the $28\times28$ pixels takes on a value in $[0,1]$.
The second corruption model is \emph{additive noise},
\begin{equation}\label{eq:additive_noise}
    \tilde{x}_{k\chi} = (1-p)x_{k\chi} + p\,\xi_{k\chi},
\end{equation}
where \(\xi_{k\chi}\) is an i.i.d.\ noise variable.
Notice that in both models, $p=0$ corresponds to the clean case and $p=1$ is the maximally corrupted case, where the training data becomes completely random.
The empirical results in the main text focus on replacement noise, while additive Gaussian noise is treated in Appendix~\ref{app:additive-gaussian}.

Although our empirical results use specific distributions (uniform and Gaussian), we show in Appendix~\ref{app:NTK} that the analytical framework applies more broadly to generic $u_{k\chi},\xi_{k\chi}$ drawn i.i.d.\ from \textit{any} distribution, since only the first two moments of the distribution enter the calculation of the signal. Equivalently, a specific choice of noise distribution can be replaced by a Gaussian with matched mean and variance without altering the result. This is similar in spirit to the Gaussian equivalence principle from the high-dimensional learning literature~\cite{mei2022generalization, goldt2022gaussian, hu2023universality, montanari2022universality}, under which the training and generalization behavior of a high-dimensional model depends on the data distribution only through its first two moments.

Importantly, the labels are assumed to remain clean throughout training. We will denote the label vector associated with class \(i=1,\ldots,C\) as \(y_{i\chi}\), with \(y_{i\chi}\) one-hot in the class index.
We assume that the dataset is class balanced, so that each of the $C$ classes contains the same number of training examples.
This assumption is not essential; it is adopted here only to streamline the discussion, and is relaxed in Appendix~\ref{app:NTK}.

\subsection{Robustness to training-set corruption}
Figure~\ref{fig:MNIST+} summarizes the basic empirical phenomenon.
Using the established image datasets MNIST, Fashion-MNIST, and KMNIST~\cite{lecun2010mnist,clanuwat2018deep,xiao2017fashion}, we show the test accuracy of trained MLPs as a function of the corruption probability \(p\) for replacement-noise corrupted training data, tested on \emph{clean} data.
Representative corrupted images are displayed below each panel.
By \(p=0.75\), the training images are already extremely difficult for a human to recognize, yet the networks still achieve strikingly high clean-test accuracy: around \(80\%\)--\(90\%\) on MNIST and \(70\%\)--\(80\%\) on Fashion-MNIST and KMNIST in this regime.
In other words, even when trained on inputs that appear almost devoid of recognizable structure, the networks retain predictive power substantially higher than random chance. In addition, we compare the empirical results to the centroid baseline on the clean data, as predicted by Eq.~\eqref{eq:maintxt_classification_rule}, finding good agreement near $p\sim0.9$. In actuality, the predicted centroid classification rule competes with a noisy background which undermines test accuracy; this will be discussed in detail in Sec.~\ref{sec:analytics}.

We also note an interesting feature of the standard deviation of the test accuracy as a function of $p$ in Fig.~\ref{fig:MNIST+}: it is non-monotonic.
Specifically, the standard deviation is relatively small near $p=1$, increases as $p$ decreases, reaches a maximum at an intermediate corruption level, and then decreases again for smaller $p$.
At first glance this is surprising: a reasonable expectation is for the network to be maximally uncertain at $p=1$, where it is trained on pure noise.
However, in Appendix~\ref{app:test_acc_var} we show that for multi-class problems ($C>2$), this is not the case: the standard deviation generically peaks at $p^{*}<1$.
This effect arises from the indicator form of the test accuracy.

\subsection{Robustness to simultaneous training-set and testing-set corruption}
\begin{figure}[t]
    \centering
    \includegraphics[width=\linewidth]{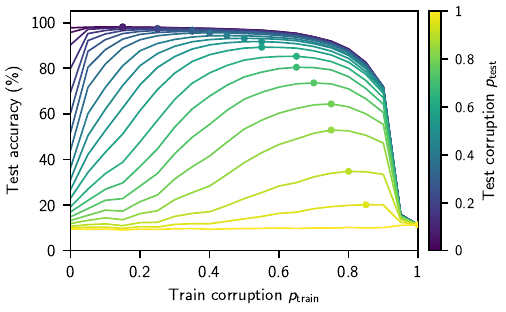}
    \caption{Train-time and test-time corruption in MNIST data: Accuracy vs.\ train corruption level $p_{\text{train}}$, for several values of test corruption $p_{\text{test}}$. We observe a non-monotonic behavior, where the optimal $p_{\text{train}}$ (dots in the figure) is near $p_{\text{test}}$, rather than 0. All networks and hyperparameters are the same as in Fig.~\ref{fig:MNIST+}.}
    \label{fig:ptrain_ptest}
\end{figure}

So far we have trained models with some corruption level $p$ and tested them on clean data.
We now consider a generalized scenario, where \emph{both} the training set and the test set are corrupted, with corruption levels $p_{\text{train}}$ and $p_{\text{test}}$, respectively.
Figure~\ref{fig:ptrain_ptest} shows the resulting test accuracy as a function of $p_{\text{train}}$, for several values of $p_{\text{test}}$ (we emphasize that the $p_{\text{test}}=0$ curve is the same as in Fig.~\ref{fig:MNIST+}).

The naive expectation might be that the optimal value of $p_{\text{train}}$ would be zero: one would expect that training with clean data will always be preferred.
However, we find that for a given value of $p_{\text{test}}$, the behavior is non-monotonic as a function of $p_{\text{train}}$, and in fact the optimal value is $p_{\text{train}}\simeq p_{\text{test}}$. The interpretation is that, when training and testing use the same level of corruption, the model is trying to generalize within the same data distribution, rather than out-of-distribution, which is more likely to succeed.
In contrast, for a fixed value of $p_{\text{train}}$, it is always optimal to use $p_{\text{test}}=0$, as is evident from Fig.~\ref{fig:ptrain_ptest}.
In Appendix~\ref{app:ptrain_ptest} we demonstrate this and also discuss a strategy to handle situations when $p_{\text{test}}$ is not known exactly.

\section{Analytical treatment}\label{sec:analytics}
We now turn to an analytical explanation of the empirical robustness observed above. We show that, in the severe-corruption regime, the network output simplifies dramatically: rather than learning a complicated denoising rule, a wide MLP trained on corrupted inputs reduces at leading order to a centroid classifier on the clean data. We derive this result using the neural tangent kernel (NTK) description of infinite-width networks, in which training with mean-squared error is equivalent to kernel regression with the NTK.

Throughout this section, we use the standard NTK initialization, with weights and biases drawn independently from mean-zero Gaussian distributions~\cite{bahri2024houches}. We also normalize both the clean training and test inputs before applying
corruption. Specifically, each input is normalized and centered so that
\begin{equation}
    \frac{1}{d}\sum_{k=1}^d x_{k\alpha}=0,
    \qquad
    \frac{1}{d}\sum_{k=1}^d x_{k\alpha}^2 = 1,
    \label{eq:main_text_rms_norm_mean_shift}
\end{equation}
where \(\alpha=\chi\) denotes a training example and \(\alpha=*\) denotes a
test example.
This normalization is applied to both the training and test sets for notational and conceptual simplicity; the more general case is discussed in Appendix~\ref{app:NTK}. Since the clean data are centered and normalized in this section, we correspondingly take the replacement noise to be
\[
u_{k\chi}\sim \mathrm{Unif}[-\sqrt{3},\sqrt{3}],
\]
so that the pure-noise inputs also have mean zero and variance one. This choice aligns the first two moments of the noise with the normalized data.

\subsection{Derivation of the high-noise centroid rule}\label{subsec:main-result}

We now present the effective decision rule learned in the high-noise regime. In the infinite-width limit, training an \(L\)-layer MLP~\footnote{we use the convention that a depth $L$ network has $L-1$ hidden layers.} with MSE loss is equivalent to kernel regression with the NTK~\cite{jacot2018ntk}. We will go through a sketch of the derivation in this section; a detailed derivation for arbitrary depth and activation is provided in Appendix~\ref{app:NTK}.

The class-\(i\) output logit of the MLP on a clean test point \(\bs{x}_*\) is given by~\cite{roberts2022principles}
\begin{equation}
    \tilde f_i(\bs{x}_*;\eps)
    =
    \tilde{\boldsymbol{\Theta}}_{*\chi}^{(L)}
    \left[\tilde{\boldsymbol{\Theta}}_{\chi\chi}^{(L)}\right]^{-1}
    \bs{y}_i,
    \label{eq:maintxt_output_general}
\end{equation}
where \(\tilde{\boldsymbol{\Theta}}_{\chi\chi}^{(L)}\) is the depth-\(L\) NTK on the noisy training set, \(\tilde{\boldsymbol{\Theta}}_{*\chi}^{(L)}\) is the test-train NTK, \(\bs{y}_i\in\{0,1\}^N\) is the target one-hot vector for class \(i\), and we have defined the signal strength
\begin{equation}
    \eps\equiv 1-p
\end{equation}
for convenience. The pure-noise limit corresponds to \(\eps=0\), while \(\eps=1\) corresponds to clean training inputs. Notice that the test-train NTK is a \(1\times N\) vector with respect to the training data, whereas the train-train NTK is an \(N\times N\) matrix in the same space.

We now sketch the derivation of the logit output in the high-noise limit. The key simplification occurs in the first layer, where the NTK has the simple form
\begin{equation}
    \tilde{\Theta}^{(1)}_{\alpha\beta}(\eps)
    =
    C_b
    +
    C_w\tilde{X}_{\alpha\beta}(\eps),
    \label{maintext_1_layer_noisy_NTK}
\end{equation}
where \(C_w>0\) and \(C_b>0\) are the weight and bias variances at initialization,
\begin{equation}
    \tilde{X}_{\alpha\beta}(\eps)
    \equiv
    \frac{1}{d}
    \sum_{k=1}^d
    \tilde{x}_{k\alpha}\tilde{x}_{k\beta}
\end{equation}
denotes the normalized noisy feature overlap matrix, and \(\alpha,\beta\) denote either train or test indices. In a mean-field-inspired decomposition, we separate the overlap matrix into a deterministic term obtained from averaging over the corruption noise, plus a subleading fluctuation
\begin{equation}
    \tilde{X}_{\alpha\beta}(\eps)
    =
    X_{\alpha\beta}(\eps)
    +
    \delta\tilde{X}_{\alpha\beta}(\eps),
    \label{maintext_overlap_mean_field}
\end{equation}
where 
\begin{equation}
    X_{\alpha\beta}(\eps)\equiv\mathbb{E}[\tilde{X}_{\alpha\beta}(\eps)]\label{eq:maintext-noise-averaged-overlap}
\end{equation}
is the noise-averaged overlap and
\begin{equation}
    \delta\tilde X_{\alpha\beta}(\eps)
    =
    \frac{1}{\sqrt d}\,
    \eta_{\alpha\beta}(\eps)
    +
    \mathcal{O}(d^{-1})
\end{equation}
is a non-deterministic fluctuation of \(\mathcal{O}(d^{-1/2})\). In the large-\(d\) limit, \(\eta_{\alpha\beta}(\eps)\) is asymptotically Gaussian with zero mean via the central limit theorem, while the \(\mathcal{O}(d^{-1})\) term captures the subleading finite-\(d\) corrections to this Gaussian limit.

Through the layer-to-layer recursion formulas~\cite{roberts2022principles,bahri2024houches}, the depth-\(L\) NTK inherits its entire data dependence, and in particular the mean-plus-fluctuation split of Eq.~\eqref{maintext_overlap_mean_field}, from \(\tilde X_{\alpha\beta}\). The full layer-wise propagation is carried out in Appendix~\ref{app:NTK}; here we simply track the two pieces of Eq.~\eqref{maintext_overlap_mean_field} through to the output.

\begin{figure*}[ht]
    \centering
    \includegraphics[width=\linewidth]{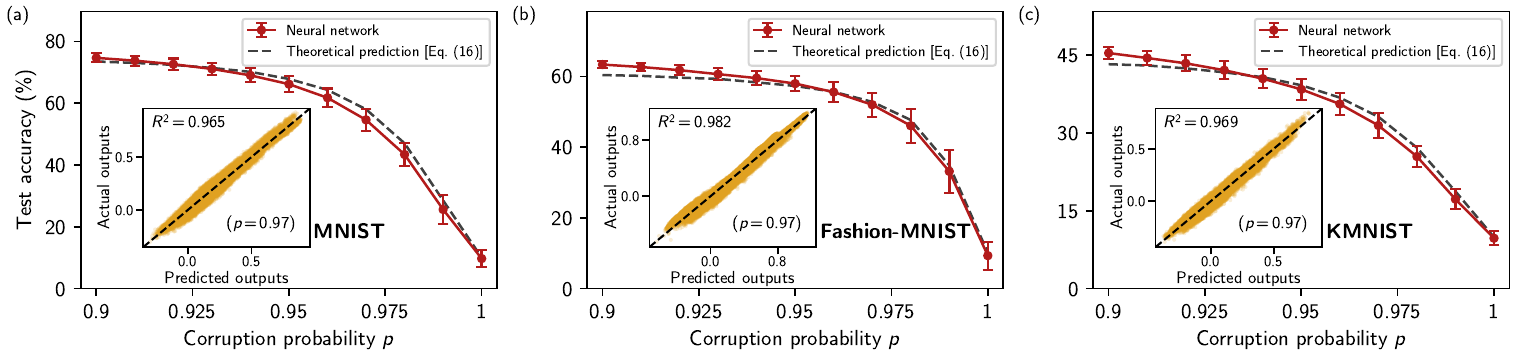}
    \caption{Numerical verification of the high-noise theory. All networks used to create the data in this figure are width-2048, 3-layer \(\mathrm{erf}\) MLPs, trained with mean squared error (MSE) loss on noisy (a)~MNIST, (b)~Fashion-MNIST, and (c)~KMNIST with \(N=4000\). Main panels: mean clean-test accuracy versus replacement-noise strength \(p\) for a nested ensemble of 20 noisy training sets and 10 MLPs per noisy set (red curve). Error bars show one standard deviation. The grey-dashed curve is the mean clean-test accuracy of the fitted effective model in Eq.~\eqref{eq:maintxt_output_fit}.
    Insets: output-level comparison at \(p=0.97\), using a nested ensemble of 100 noisy training sets and 50 MLPs per noisy set. Each point corresponds to one clean test image and one candidate class, comparing the ensemble-mean network output with the fitted prediction from Eq.~\eqref{eq:maintxt_output_fit}. The black dashed line in the inset indicates perfect agreement.}
    \label{fig:actual_vs_predicted}
\end{figure*}

We work near the pure-noise point and in the large-feature-dimension limit: \(\eps\ll1\) and \(\sqrt{d}\gg1\). Throughout this expansion, the argument \((0)\) denotes evaluation at pure noise, \(\eps=0\) (equivalently \(p=1\)). With this in mind, we perform a joint first-order expansion of the noisy overlap matrix Eq.~\eqref{maintext_overlap_mean_field} in \(\eps\) and \(1/\sqrt{d}\), leading to
\begin{equation}
\begin{split}
    \tilde{X}_{\alpha\beta}(\eps)
    &=
    X_{\alpha\beta}(0)
    +
    \eps\,
    \partial_\eps X_{\alpha\beta}(0)
    +
    \frac{1}{\sqrt{d}}\eta_{\alpha\beta}(0)
    \\
    &\quad
    +
    \mathcal{O}(\eps^2)
    +
    \mathcal{O}(\eps d^{-1/2})
    +
    \mathcal{O}(d^{-1}),
    \label{maintext_overlap_expansion}
\end{split}
\end{equation}
where we have also expanded the non-deterministic fluctuation \(\eta_{\alpha\beta}\) about \(\eps=0\).

Feeding this structure through the recursion, the depth-\(L\) NTK takes on a similar form,
\begin{align}
    \tilde \Theta_{\alpha\beta}^{(L)}(\eps)
    &=
    \Theta_{\alpha\beta}^{(L)}(0)
    +
    \eps
    \partial_\eps\Theta_{\alpha\beta}^{(L)}(0)
    +
    \frac{1}{\sqrt d}\eta^{(L)}_{\alpha\beta}(0)
    \notag\\
    &+
    \mathcal{O}(\eps^2)
    +
    \mathcal{O}(\eps d^{-1/2})
    +
    \mathcal{O}(d^{-1}),
    \label{eq:maintxt_first_layer_expansion}
\end{align}
where \(\Theta_{\alpha\beta}^{(L)}(0)\) is the depth-\(L\) NTK built from the pure-noise averaged overlap \(X_{\alpha\beta}(0)\), and \(\eta^{(L)}_{\alpha\beta}(0)\) is zero-mean Gaussian fluctuation inherited from \(\delta\tilde X\).

Substituting Eq.~\eqref{eq:maintxt_first_layer_expansion} into Eq.~\eqref{eq:maintxt_output_general} and expanding to leading order in \(\eps\) and \(d^{-1/2}\), we obtain
\begin{align}
    f_i(\bs{x}_*;\eps)
    &\approx
    C^{(L)}
    +
    A^{(L)}
    \sqrt{\frac{N/C}{d}}\,
    \hat\xi_i
    \notag\\
    &+
    B^{(L)}
    \frac{\eps N}{C}
    \left[
        \frac{1}{d}
        \sum_{k=1}^d
        x_{k*}\bigl(\bar x_k^i-\bar x_k\bigr)
    \right],
    \label{eq:maintxt_final_gaussian_output}
\end{align}
where \(A^{(L)}\), \(B^{(L)}\), and $C^{(L)}$ are class-independent constants which encode depth, activation, initialization, and details of the noise distribution; \(\hat\xi_i\) is a class-dependent Gaussian fluctuation with zero mean and unit variance, and
\begin{equation}
    \bar x_k^i
    =
    \frac{C}{N}\sum_{\chi\in i}x_{k\chi},
    \qquad
    \bar x_k
    =
    \frac{1}{N}\sum_{\chi}x_{k\chi}
    \label{eq:maintxt_centroid}
\end{equation}
are the centered empirical mean, or centroid, of class \(i\) and the global empirical mean, respectively, in feature \(k\) for the \emph{clean} training data. The crucial point is that the leading class-dependent signal depends on the training data only through these \emph{clean} centered-class means. Despite being trained on corrupted inputs, the network extracts from the noisy data a class-dependent statistic depending only on the clean dataset. At this order, the noisy part of the inputs contributes only the fluctuation term \(\hat\xi_i\), while the discriminating signal is the overlap between the test point and the clean class centroid.

Therefore, the leading-order decision rule learned by the network in the joint limit $\eps\ll1$ and $\sqrt{d}\gg1$ is
\begin{equation}
    i^\star
    =
    \arg\max_{i=1,\ldots,C}
    \frac{1}{d}
    \sum_{k=1}^d
    x_{k*}\bigl(\bar x_k^i-\bar x_k\bigr).
    \label{eq:maintxt_classification_rule}
\end{equation}
In other words, the NTK calculation predicts that, in the high-noise regime, the network reduces to a nearest-class-mean, or centroid, classifier~\cite{tibshirani2002diagnosis} with an inner-product score in feature space. Remarkably, the leading-order decision rule is universal across depth, activation, and noise distribution. Despite the depth and nonlinearity of the original model, classification collapses to simple linear template matching against one prototype per class.

We note that the coefficient \(B^{(L)}\) multiplying the centroid term in Eq.~\eqref{eq:maintxt_final_gaussian_output} is guaranteed to be nonnegative under our normalization convention Eq.~\eqref{eq:main_text_rms_norm_mean_shift} and unit-variance noise models; see Appendix~\ref{app:B_nonnegative} for further details. However, nonnegativity of \(B^{(L)}\) is not guaranteed for arbitrary non-normalized data or completely general noise models. For settings where \(B^{(L)}<0\), the leading decision rule is simply obtained by reversing the orientation of the centroid score, equivalently by replacing \(\arg\max\) with \(\arg\min\). Thus, the leading class-dependent response remains a centroid-overlap rule, with the sign of \(B^{(L)}\) determining its orientation.

Finally, we comment on the controllability of the approximation at large training dataset size \(N\). The prefactor of the centroid score in Eq.~\eqref{eq:maintxt_final_gaussian_output} comes from expanding the inverse NTK around the pure-noise kernel. For moderate \(N\), this expansion is controlled by the small signal strength \(\eps\). At larger \(N\), higher-order terms in \(\eps\) can become important; however, these terms primarily renormalize the overall coefficient of the centroid statistic, leaving the leading decision rule unchanged (further details are given in Appendix~\ref{app:NTK}). The full expansion obtained by substituting higher-order NTK corrections into Eq.~\eqref{eq:maintxt_output_general} does not necessarily converge uniformly in \(N\). Nonetheless, as shown in the next subsection, the leading correction already agrees well with the empirical network outputs and test accuracy in the high-noise regime.

\subsection{Comparison with finite-width networks}

We now compare the theoretical prediction for the logit output, Eq.~\eqref{eq:maintxt_final_gaussian_output}, to finite-width networks in a controlled empirical setting matched to the assumptions of the high-noise NTK calculation. This setting differs from the motivating experiments in Fig.~\ref{fig:MNIST+}, which use finite-width ReLU MLPs, cross-entropy loss, and raw image inputs in $[0,1]$ to establish the robustness phenomenon in a standard classification setup. Here, by contrast, we use the normalization convention of Eq.~\eqref{eq:main_text_rms_norm_mean_shift} for the inputs, MSE loss, wider networks with erf activation, and fewer training data points so that the numerical experiment isolates the leading-order mechanism predicted by the theory. The theoretical prediction is not intended to reproduce every implementation detail of Fig.~\ref{fig:MNIST+}; rather, it identifies a tractable high-noise mechanism, which we validate directly in the finite-width experiments of Fig.~\ref{fig:actual_vs_predicted}. Further experimental details on the empirical results presented here can be found in Appendix~\ref{app:experiment-details}.

Restoring the corruption probability \(p=1-\eps\), the form of Eq.~\eqref{eq:maintxt_final_gaussian_output} motivates the effective model
\begin{equation}
    f_i(\bs{x}_*)
    =
    a(1-p)
    \left[
        \frac{1}{d}
        \sum_{k=1}^d x_{k*}\bigl(\bar x_k^i-\bar x_k\bigr)
    \right]
    +
    b
    +
    c\hat\xi_i,
    \label{eq:maintxt_output_fit}
\end{equation}
where \(a\), \(b\), and \(c\) absorb the renormalized architecture-dependent prefactors and fluctuation scale. For simplicity, we model the fluctuations \(\hat{\xi}_i\) as i.i.d.\ standard normal random variables. To estimate these coefficients, we train \(m\) MLPs for each of \(M\) independent noisy-dataset realizations, for a total of \(mM\) trained networks. For each test point and class, we first average the output logit over the \(m\) networks corresponding to a fixed noisy realization. This produces \(M\) realization-averaged logits per class, which are the empirical objects to be compared with Eq.~\eqref{eq:maintxt_output_fit}. We then average these logits over the \(M\) noisy realizations and determine \(a\) and \(b\) by a least-squares fit over all test points and classes to the mean of Eq.~\eqref{eq:maintxt_output_fit}; the coefficient \(c\) does not enter this step because it multiplies a zero-mean fluctuation. Finally, we set \(c\) equal to the standard deviation of the \(M\) realization-averaged logits, pooled over all test points and classes. Importantly, this is a global three-parameter fit: we do \emph{not} fit separate coefficients for different test examples. 

Figure~\ref{fig:actual_vs_predicted} tests this picture in two complementary ways. The insets compare the empirical mean output logits of the MLPs on the full MNIST, Fashion-MNIST, and KMNIST test sets with the prediction obtained from the mean of Eq.~\eqref{eq:maintxt_output_fit}. After fitting the two global coefficients \(a\) and \(b\), we find nearly exact agreement for each dataset. This provides strong evidence that, after averaging over noise realizations, the outputs of the MLP collapse onto the mean of the centroid model, consistent with centroid-classifier behavior. The main panels compare the clean-test accuracy of the empirical ensemble with that of the full centroid model in Eq.~\eqref{eq:maintxt_output_fit}, now including the fluctuation term controlled by the fitted coefficient \(c\). The agreement for each dataset shows that the centroid model captures not only the mean output structure, but also the fluctuations that govern classification accuracy. In particular, it accurately reproduces the decay of the test accuracy as the noise probability \(p\) increases.
The main panels also indicate where the empirical networks begin to depart from the leading-order centroid description, signaling the onset of higher-order corrections in \((1-p)\). In our data, this deviation becomes noticeable for \(p \lesssim 0.9\). \ We provide additional tests in Appendix \ref{app:test_acc_p=0-1} and \ref{app:architectural_universality}.

The agreement in Fig.~\ref{fig:actual_vs_predicted} should therefore be interpreted as a controlled validation of the high-noise centroid mechanism, rather than as a claim that the theory reproduces every implementation detail of the raw-input cross-entropy experiments in Fig.~\ref{fig:MNIST+}. Nevertheless, it indicates that the same centroid mechanism remains visible in finite-width networks once the experimental setting is chosen to match the assumptions of the calculation.

\subsection{Signal-to-noise ratio}

\begin{figure*}[th]
    \centering
    \includegraphics[width=\linewidth]{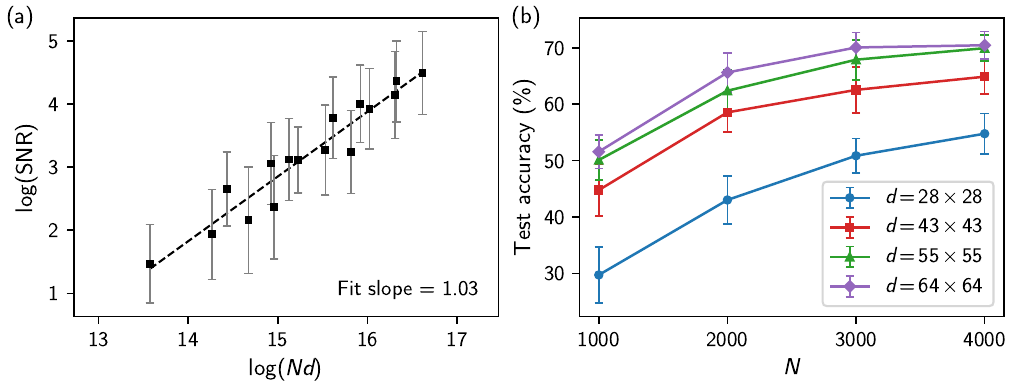}
    \caption{
    Empirical $N,d$ scaling at fixed replacement-noise probability $p=0.97$ and number of classes $C=10$ for width-2048, 3-layer $\mathrm{erf}$ MLPs trained with MSE loss on MNIST.
    (a)~Mean $\log(\mathrm{SNR})$ versus $\log(Nd)$ over 500 randomly sampled clean-test examples. For each example, true-class logits are first averaged over 10 independently initialized networks trained on each noisy training dataset. The SNR is the squared mean of these averaged logits divided by their sample variance across 20 noisy training datasets. Squares and error bars show the mean and one standard deviation of log(SNR) across test examples; the dashed line is a linear fit to the means.
    (b)~Mean clean-test accuracy versus $N$ at fixed $d$, using predictions from initialization-averaged logits. Means and one-standard-deviation error bars are computed across the 20 noisy training datasets. Larger feature dimensions $d$ are obtained by resizing MNIST images using bilinear interpolation.}
    \label{fig:scaling_law}
\end{figure*}

Equation~\eqref{eq:maintxt_final_gaussian_output} for the output logit also implies a scaling law, since it predicts that, at least for intermediate $N$, the class-dependent mean grows as \((1-p)(N/C)\) while the  standard deviation scales as \(\sqrt{(N/C)/d}\). Therefore, we expect the signal-to-noise ratio (SNR) to scale as
\begin{equation}
    \mathrm{SNR}
    \sim
    (1-p)^2d\frac{N}{C}
    \label{eq:maintxt_snr_scaling}
\end{equation}
The theory therefore predicts improved classification accuracy as the surviving signal fraction \(1-p\), the feature dimension \(d\), or the number of training examples per class \(N/C\) are increased.
We test the $N,d$ dependence on MNIST at fixed replacement-noise probability $p=0.97$ and class count $C=10$. With these quantities held fixed, Eq.~\eqref{eq:maintxt_snr_scaling} predicts $\mathrm{SNR}\propto Nd$, corresponding to a unit slope on logarithmic axes. Figure~\figref{fig:scaling_law}{a} directly tests this prediction using the empirical SNR of the true-class output logit. Averaging $\log(\mathrm{SNR})$ over randomly sampled clean-test examples, the results for different $N$ and $d$ approximately collapse onto a linear trend in $\log(Nd)$. The fitted slope of $1.03$ is close to the predicted value of unity, supporting the proposed scaling. Figure~\figref{fig:scaling_law}{b} shows the accompanying improvement in clean-test accuracy with increasing $N$ and $d$.

The analysis in this section has focused on uniform replacement noise for concreteness.
However, the centroid-classification result is not tied to that specific choice of corruption model.
Rather, for the two broad classes of input noise most relevant in practice---general replacement noise and general additive noise---the same leading centroid structure persists.
In Appendix~\ref{app:additive-gaussian} we present analogous results for the additive noise model [Eq.~\eqref{eq:additive_noise}], demonstrating the generality of our findings.

\subsection{Permutation symmetry and the centroid response}
\label{subsec:symmetry-origin}

The centroid rule Eq.~\eqref{eq:maintxt_classification_rule} is independent of noise distribution, depth, and activation. Here we give intuition for this universality within the NTK framework using a linear-response argument: we treat the trained network as a system sitting at a permutation-symmetric pure-noise reference state and view the surviving signal as a small perturbation away from it. This provides a symmetry interpretation of the high-noise expansion derived in Appendix~\ref{app:NTK}. It should be read as a minimal linear-response model rather than a second microscopic derivation of the corruption channels of Sec.~\ref{sec:numerics}, but it is sufficient to extract the leading-order class-dependent decision rule Eq.~\eqref{eq:maintxt_classification_rule}. 

After separating the $\mathcal{O}(d^{-1/2})$ stochastic fluctuations, the deterministic part of the noisy NTK depends on the data only through the noise-averaged overlap Eq.~\eqref{eq:maintext-noise-averaged-overlap}. At pure noise, $\eps=1-p=0$, the training examples become exchangeable: the noise-averaged train-train overlap is invariant under any relabeling $\pi\in S_N$ of the training examples,
\begin{equation}
    X_{\pi(\alpha)\pi(\beta)}(0) = X_{\alpha\beta}(0).
    \label{eq:maintext_perm_invariance}
\end{equation}
That is, $X_{\alpha\beta}(0)$ depends on $\alpha,\beta$ only through whether $\alpha=\beta$. This symmetry structure is guaranteed at pure noise for any distribution that is i.i.d.\ with respect to data and feature space and has bounded mean and variance.

Motivated by this structure, we introduce a minimal pure-noise reference ensemble for the training set. This reference ensemble consists of featureless, i.i.d.\ random variables $\{\xi_{k\alpha}\}$, whose purpose is to capture the exchangeable structure of the deterministic part of the noisy NTK at the pure-noise fixed point. The corresponding pure-noise-averaged test--train and train--train overlap matrices are
\begin{align}
    X_{0,*\alpha}
    &=
    \frac{1}{d}\sum_{k=1}^d x_{k*}\mathbb{E}\!\left[\xi_{k\alpha}\right],
    \label{eq:maintext_test_train_overlap}
    \\
    X_{0,\alpha\beta}
    &=
    \frac{1}{d}\sum_{k=1}^d
    \mathbb{E}\!\left[\xi_{k\alpha}\xi_{k\beta}\right],
    \label{eq:maintext_train_train_overlap}
\end{align}
where the expectation $\mathbb{E}[\cdot]$ is taken with respect to the noise distribution. We leave the first two moments of the reference noise arbitrary, assuming only that they are bounded. For later convenience, we assume the first moment vanishes and denote the second moment by $v$~\footnote{The value of the first moment is inconsequential for the rest of the calculation; a nonzero first moment simply renormalizes the constants $r^{(\ell)}$, $s^{(\ell)}$, and $t^{(\ell)}$ introduced later in the subsection. In contrast, the value of the second moment can affect the orientation of the centroid rule, as discussed below.}. Adopting the normalization convention of Eq.~\eqref{eq:main_text_rms_norm_mean_shift} for the clean training inputs $\{x_{k\alpha}\}$ and test input $x_{k*}$, the imposed mean-centering makes the pure-noise test--train overlap vanish, $X_{0,*\alpha}=0$, while the pure-noise train--train overlap remains diagonal $X_{0,\alpha\beta}=\delta_{\alpha\beta}v$.

We treat the network trained on the pure-noise ensemble $\{\xi_{k\alpha}\}$ as a pure-noise fixed point. Now suppose we perturb the pure-noise inputs by a small signal proportional to the clean features,
\begin{equation}
    \xi_{k\alpha}
    \to
    \xi_{k\alpha}+\eps\,x_{k\alpha}.
    \label{eq:maintext_perturb_fixed_point}
\end{equation}
To study the response, let \(\delta\) denote the leading-order variation along the perturbation path
\(z_{k\alpha}(\eps)=\xi_{k\alpha}+\eps x_{k\alpha}\). That is, for any function \(F\) built from the training inputs, define
\begin{equation}
    \delta F
    \equiv
    \left.
    \eps\frac{d}{d\eps}\right|_{\eps=0}
    F\bigl[\bs{z}(\eps)\bigr],
\end{equation}
so that \(\delta F=\mathcal{O}(\eps)\). The leading-order response of the test--train overlap is
\begin{equation}
    \delta X_{0,*\alpha}
    =
    \frac{\eps}{d}
    \sum_{k=1}^d x_{k*}x_{k\alpha},
    \label{eq:maintext_test_train_overlap_variation}
\end{equation}
while the response of the train--train overlap vanishes, since its leading nonzero term is $\mathcal{O}(\eps^2)$. Thus only $\delta X_{0,*\alpha}$ carries the signal at leading order~\footnote{The train--train overlap vanishes at linear order in $\eps$ under our convention that each training input has zero mean across features; see Eq.~\eqref{eq:main_text_rms_norm_mean_shift}.}.

We now compute the response at arbitrary depth $L$. Let $\Theta^{(\ell)}_{0,\alpha\beta}$ and $\Theta^{(\ell)}_{0,*\alpha}$ denote the depth-$\ell$ train--train and test--train NTKs at pure noise, respectively. At leading order in $\eps$, the class-$i$ logit responds as
\begin{equation}
    \delta f_i(\bs{x}_*)
    =
    \sum_{\alpha,\beta=1}^N
    \delta\Theta^{(L)}_{0,*\alpha}
    \left[\bs{\Theta}^{(L)}_0\right]_{\alpha\beta}^{-1}
    y_{i\beta},
    \label{eq:maintext_logit_response}
\end{equation}
where the variation of the inverse train--train NTK vanishes at leading order in this minimal treatment. Two objects therefore determine the response: the inverse pure-noise train--train NTK $[\bs{\Theta}^{(L)}_0]^{-1}$ and the pure-noise test--train NTK variation $\delta\Theta^{(L)}_{0,*\alpha}$. Both follow from applying the layer-to-layer recursion relations~\cite{roberts2022principles} to the pure-noise fixed point:
\begin{align}
    K^{(\ell+1)}_{0,\alpha\beta}
    &=
    C_b + C_w\langle\sigma_\alpha\sigma_\beta\rangle_{\bs{K}_0^{(\ell)}},
    \label{eq:maintext_preact_recursion}
    \\
    \Theta^{(\ell+1)}_{0,\alpha\beta}
    &=
    K^{(\ell+1)}_{0,\alpha\beta}
    +
    C_w\langle\sigma'_\alpha\sigma'_\beta\rangle_{\bs{K}_0^{(\ell)}}
    \Theta^{(\ell)}_{0,\alpha\beta},
    \label{eq:maintext_ntk_recursion}
\end{align}
where $\bs{K}^{(\ell)}_0$ denotes the depth-$\ell$ pure-noise pre-activation kernel, and the layer-$1$ initial condition is fixed directly by the pure-noise overlap,
\begin{equation}
    K^{(1)}_{0,\alpha\beta}
    =
    \Theta^{(1)}_{0,\alpha\beta}
    =
    C_b+C_wX_{0,\alpha\beta}.
\end{equation}
In the recursion relations, $\sigma_\alpha\equiv\sigma(u_\alpha)$, $\sigma'_\alpha$ is its derivative, and $\langle\cdots\rangle_{\bs{K}}$ is an expectation over the zero-mean bivariate Gaussian pre-activations $(u_\alpha,u_\beta)$ with $2\times2$ covariance $\bs{K}$ [Eq.~\eqref{eq:app_2d_gaussian_expectation}].

Consider first the inverse train--train NTK. At pure noise the train--train overlap is permutation-invariant, and hence so is the layer-$1$ NTK, $\Theta^{(1)}_{0,\alpha\beta}=r^{(1)}\delta_{\alpha\beta}+s^{(1)}$, where $r^{(1)}=C_wv$ and $s^{(1)}=C_b$. The recursions at pure-noise are permutation-equivariant; if the depth-$1$ kernel is permutation invariant, then this invariance is preserved at arbitrary depth $\ell$. Therefore, the depth-$L$ train--train NTK therefore retains the two-parameter form
\(\Theta^{(L)}_{0,\alpha\beta}=r^{(L)}\delta_{\alpha\beta}+s^{(L)}\),
for constants $r^{(L)}$ and $s^{(L)}$ that depend on depth, activation, initialization, and the moments of the noise distribution. Equivalently, the depth-$L$ NTK at pure noise lives in the projector algebra $\mathrm{span}(\bs{P}_\parallel,\bs{P}_\perp)$, with $\bs{P}_\parallel=\frac{1}{N}\bs{1}\bs{1}^{\mathrm{T}}$ the projector onto the permutation-invariant sector and $\bs{P}_\perp=\bs{I}-\bs{P}_\parallel$ its complement:
\begin{equation}
    \bs{\Theta}^{(L)}_0
    =
    r^{(L)}\bs{P}_\perp
    +
    \left(r^{(L)}+s^{(L)}N\right)\bs{P}_\parallel.
    \label{eq:maintext_projector}
\end{equation}
The NTK is an example of a Gram matrix, which is strictly positive-definite for distinct inputs~\cite{jacot2018ntk}. Therefore, its eigenvalues $r^{(L)}>0$ and $r^{(L)}+s^{(L)}N>0$ are strictly positive, and so it has a well-defined inverse
\begin{equation}
    \left[\bs{\Theta}^{(L)}_0\right]^{-1}
    =
    \frac{\bs{P}_\perp}{r^{(L)}}
    +
    \frac{\bs{P}_\parallel}{r^{(L)}+s^{(L)}N}.
    \label{eq:maintext_inverse_NTK}
\end{equation}

Consider next the depth-$L$ test--train NTK variation $\delta\Theta_{0,*\alpha}^{(L)}$. The first-layer variation about pure noise is
\begin{equation}
    \delta\Theta^{(1)}_{0,*\alpha}
    =
    \delta K^{(1)}_{0,*\alpha}
    =
    C_w\,\delta X_{0,*\alpha}.
\end{equation}
Layer-to-layer propagation of the variation can be handled by differentiating the recursion relations using Price's theorem~\cite{price2003useful}. The resulting recursion relations for the test--train kernel and NTK variations about pure noise are
\begin{align}
    \delta K_{0,*\alpha}^{(\ell+1)}
    &=
    C_w\langle\sigma'_*\sigma'_\alpha\rangle_{\bs{K}_0^{(\ell)}}
    \,\delta K_{0,*\alpha}^{(\ell)},
    \label{eq:maintext_kernel_variation_recursion}
    \\
    \delta\Theta_{0,*\alpha}^{(\ell+1)}
    &=
    \delta K_{0,*\alpha}^{(\ell+1)}
    +
    C_w\langle\sigma''_*\sigma''_\alpha\rangle_{\bs{K}_0^{(\ell)}}
    \Theta^{(\ell)}_{0,*\alpha}
    \delta K_{0,*\alpha}^{(\ell)}
    \nonumber\\
    &\quad
    +
    C_w\langle\sigma'_*\sigma'_\alpha\rangle_{\bs{K}_0^{(\ell)}}
    \,\delta\Theta^{(\ell)}_{0,*\alpha},
    \label{eq:maintext_NTK_variation_recursion}
\end{align}
where we have used the fact that diagonal variations of the test--train kernel and NTK vanish at leading order. 

We now show how permutation symmetry in the training index $\alpha$ reduces layer and activation dependence to an overall scaling of the first-layer variation. To see this, consider the entries of the layer-$1$ test--train kernel at pure-noise, which are permutation-invariant in training input $\alpha$
\begin{equation}
    K^{(1)}_{0,**}=C_b+C_w,\quad
    K^{(1)}_{0,\alpha\alpha}=C_b+C_wv,\quad
    K^{(1)}_{0,*\alpha}=C_b,
    \label{eq:maintext_layer1_testtrain_entries}
\end{equation}
where $v$ is the second moment of the noise distribution. Permutation equivariance of the pure-noise recursion Eq.~\eqref{eq:maintext_preact_recursion}
preserves this $\alpha$-independence to arbitrary depth. Consequently, the pure-noise
activation-dependent expectation values and NTK entries appearing in Eqs.~\eqref{eq:maintext_kernel_variation_recursion}
and~\eqref{eq:maintext_NTK_variation_recursion} are independent of $\alpha$. Since $\delta\Theta^{(1)}_{0,*\alpha}\propto\delta X_{0,*\alpha}$, we conclude that the recursion merely rescales the pure-noise overlap variation by a constant:
\begin{equation}
    \delta\Theta^{(L)}_{0,*\alpha}
    =
    t^{(L)}\,\delta X_{0,*\alpha},
    \label{eq:maintext_ntk_variation}
\end{equation}
where $t^{(L)}$ depends on depth, activation, initialization, and the moments of the noise distribution.

Substituting Eqs.~\eqref{eq:maintext_inverse_NTK} and \eqref{eq:maintext_ntk_variation} into Eq.~\eqref{eq:maintext_logit_response}, and using the class-balanced label contractions
\begin{equation}
    \bs{P}_\perp\bs{y}_i
    =
    \bs{y}_i-\frac{1}{C}\bs{1},
    \qquad
    \bs{P}_\parallel\bs{y}_i
    =
    \frac{1}{C}\bs{1},
    \label{eq:maintext_label_contractions}
\end{equation}
the permutation-invariant sector $\bs{P}_\parallel$ contributes only a class-independent constant, while the $\bs{P}_\perp$ sector carries the class signal. With $\delta X_{0,*\alpha}$ from Eq.~\eqref{eq:maintext_test_train_overlap_variation}, this gives
\begin{equation}
\begin{split}
    \delta f_i(\bs{x}_*)
    &=
    \frac{t^{(L)}}{r^{(L)}}
    \frac{\eps N}{C}
    \left[
        \frac{1}{d}
        \sum_{k=1}^d
        x_{k*}
        \left(\bar{x}_k^i-\bar{x}_k\right)
    \right]
    \\
    &\quad
    + \text{class-independent terms},
    \label{eq:maintext_centroid_response}
\end{split}
\end{equation}
which is exactly the centroid rule. The constants $t^{(L)}$ and $r^{(L)}$ encode depth, activation, initialization, and the first two moments of the noise distribution, while the class-dependent structure is fixed by permutation symmetry at the pure-noise fixed point.

A remark on orientation is useful. The prefactor $t^{(L)}/r^{(L)}$ is common to all classes, so only its sign matters: a positive sign gives the nearest-class-mean rule, while a negative sign reverses the orientation of the centroid score. Since $r^{(L)}>0$, the orientation is determined by the sign of $t^{(L)}$. Under the unit-variance convention $v=1$, the layer-$1$ test--train kernel entries become
\begin{equation}
    K^{(1)}_{0,**}
    =
    \left.K^{(1)}_{0,\alpha\alpha}\right|_{v=1}
    =
    C_b+C_w,
    \qquad
    K^{(1)}_{0,*\alpha}
    =
    C_b.
\end{equation}
which can be determined using Eq.~\eqref{eq:maintext_layer1_testtrain_entries}. Thus the test and training diagonals coincide, and the off-diagonal entry is nonnegative. These properties imply that the activation-dependent transfer factors \(\langle h_*h_\alpha\rangle_{\bs{K}_0^{(\ell)}}\), with \(h\) being any of \(\sigma,\sigma',\sigma''\), appearing in the kernel and NTK variation recursions are nonnegative at every depth for any activation function---we show this in detail in Appendix~\ref{app:B_nonnegative}. Consequently, for unit-variance noise under the normalization convention Eq.~\eqref{eq:main_text_rms_norm_mean_shift}, we have \(t^{(L)}\geq 0\).

For general $v\neq1$ or non-normalized data, the test and training diagonals of the pure-noise test--train kernel need not coincide, and the sign is not guaranteed for every activation. If the prefactor is negative, however, the leading decision rule is obtained by reversing the orientation of the centroid score, equivalently by replacing $\arg\max$ with $\arg\min$. Thus the leading class-dependent response remains a centroid-overlap rule for any depth, activation, and i.i.d.\ noise distribution, with the sign of $t^{(L)}$ fixing the orientation.

\section{Discussion and outlook}\label{sec:discussion}

The problem of training neural networks with heavily corrupted inputs has revealed numerous surprises.
First, we found that MLPs are remarkably robust to attribute noise, capable of classifying heavily corrupted images with accuracy much higher than chance.
This phenomenon, observed across several datasets, prompted us to tackle the problem analytically, with the naive expectation of revealing the complicated ``denoising'' process employed by the networks, enabling them to perform so well.
Using a controlled expansion of the NTK in the strong corruption regime, we found that in fact the networks learn a very simple rule: they compare each test image to the average of all training images of a certain class, and check which one is the most similar.
This nearest-class-mean (centroid) rule, which is the dominant contribution in the strong corruption limit, emerges despite the complexity and nonlinearity enabled by the network.
The analytical result also leads to an effective signal-to-noise ratio [Eq.~\eqref{eq:maintxt_snr_scaling}], which reveals that performance improves with feature dimension and dataset size; we have confirmed that empirically as well.
Taken together, these results show that, in the high-attribute-noise regime, wide MLPs have a universal leading class-dependent response: a centered centroid classifier. The finite-width networks studied here visibly follow this mechanism, explaining their surprising robustness on the MNIST-family datasets.

The symmetry interpretation in Sec.~\ref{subsec:symmetry-origin} explains why this simple centroid rule is not a coincidence of the first-layer calculation: it is fixed by the exchangeability of the pure-noise kernel and by the permutation equivariance of the NTK recursion.
This observation also suggests a broader interpretation of the formalism.
The essential ingredients are not the particular MLP recursion alone; the same reasoning should apply beyond fully connected MLPs to any architecture with a well-defined kernel limit whose pure-noise kernel has the same exchangeability structure.
When those ingredients are present, linear response around the high-noise fixed point identifies the first class-dependent correction to the predictor; in the present setting, that correction is precisely the centroid rule.
More broadly, this points to a way of understanding learning phenomena by locating tractable fixed points of a network's effective description and deriving the leading decision-rule corrections that appear when a small signal breaks the fixed-point symmetry.

The success of the NTK prediction for finite-width networks suggests that the high-noise regime has an effective universality beyond the exactly solvable infinite-width limit.
Although finite width, nonlinear feature learning, and the choice of loss can change architecture-dependent prefactors and fluctuation scales, the dominant class-dependent direction appears to remain the centered centroid overlap.
This should not be read as a claim that MLPs are robust to arbitrary training corruption on every classification dataset.
Rather, the theory predicts a generic leading-order decision rule for MLPs in the high-noise regime, while the resulting accuracy depends on the geometry of the dataset itself.
In the datasets studied here, MNIST, Fashion-MNIST, and KMNIST, the clean class centroids contain enough discriminative information to yield well-above-chance classification, which explains why the networks remain robust at high corruption in Fig.~\ref{fig:MNIST+}.

Finally, the analytical control enabled in the strong-corruption regime may also have broader implications.
Our perturbative result in $(1-p)$ could, in principle, be extended to higher orders, thereby revealing progressively finer structure in the rules learned by the network.
Systematically carrying out this expansion may provide access to the intermediate-corruption regime, or even suggest a controlled route toward the clean-data limit, using the corruption strength as an explicit expansion parameter.
A related possible direction is the analysis of denoising models, including diffusion-type models, in the high-noise regime.
In that setting, one would replace the classification target considered here by a regression target, such as the clean input.
An analogous expansion around the pure-noise limit could then reveal what statistics of the clean data are recovered at leading order, and how progressively finer denoising information emerges at higher orders in the signal strength.

\section*{Acknowledgments}
We thank P.\ Frazier, T.\ Arias, C.-M.\ Jian, H.\ Pan, T.\ Nebabu, and J.\ Kim for useful discussions.
J.T.\ acknowledges funding from the Center for Alkaline-Based Energy Solutions (CABES), part of the Energy Frontier Research Center (EFRC) program supported by the U.S. Department of Energy, under grant DE-SC-0019445. A.B.\ was supported by a Cornell fellowship from the Cornell University Graduate School. H.K.\ acknowledges support by the NSF through the grant OAC-2118310. O.L.\ acknowledges support from the Bethe-KIC postdoctoral fellowship at Cornell University.

\bibliography{library}

\appendix
\onecolumngrid 
\newpage
\section{Details of the experiments}\label{app:experiment-details}
\begin{table*}[!htbp]
\label{tab:experiment-details}
\centering
\footnotesize
\setlength{\tabcolsep}{5pt}
\renewcommand{\arraystretch}{1.2}
\begin{tabular}{p{0.10\textwidth}p{0.36\textwidth}p{0.44\textwidth}}
\toprule
Figure & Dataset and corruption & Model and training \\
\midrule

Fig.~\ref{fig:MNIST+}
& Balanced MNIST, Fashion-MNIST, KMNIST; 54,210 training and 8,920 clean test images each; replacement noise with $u\sim{\rm Unif}[0,1]$, fixed across epochs; empirical $[0,1]$ inputs with no per-example mean-shift or unit-RMS normalization
& Width-128, 3-hidden-layer ReLU MLP; cross-entropy; Adam, learning rate $10^{-3}$; 20 epochs; batch size 128; 100 noisy training realizations per $p$ on a 21-point grid, denser above $0.8$. Mean final-model accuracy with population SD. Horizontal clean-centroid references use inner-product classification on per-image centered, unit-RMS clean training and test data; no $p$ dependence \\
\addlinespace

Fig.~\ref{fig:ptrain_ptest}
& MNIST; full 60,000 training and 10,000 test images; independent train/test replacement noise, $u\sim{\rm Unif}[0,1]$; raw $[0,1]$ inputs. Training noise is resampled on each image access; one fixed test corruption per $p_{\rm test}$ is shared across runs
& Same architecture, optimizer, epoch budget, batch size, and raw-input convention as Fig.~\ref{fig:MNIST+}; 20 training repeats per $p_{\rm train}$; $p_{\rm train}=0,0.05,\ldots,1$ and $p_{\rm test}=0,0.0125,\ldots,1$. Mean final-model accuracy; same experiment as Fig.~\ref{fig:sbs_ptrain_ptest_auc} \\
\addlinespace

Fig.~\ref{fig:actual_vs_predicted}
& MNIST, Fashion-MNIST, KMNIST; balanced $N=4000$, $d=784$, and 10,000 clean test images each; replacement noise at $p=0.90,\ldots,1.00$ in steps of $0.01$. Clean train/test vectors are mean-centered and scaled to centered RMS 1 before corruption; $u\sim{\rm Unif}[-\sqrt{3},\sqrt{3}]$; no post-corruption renormalization; noise fixed across epochs
& Width-2048, 3-layer $(L=3)$ erf MLP, counting the output layer; MSE loss; Adam, learning rate $10^{-4}$; 30 epochs; batch size 250. Accuracy: 20 noisy datasets, 10 MLPs each; average logits within each dataset, then report mean accuracy and sample SD across datasets. Inset and calibration at $p=0.97$: 100 noisy datasets, 50 MLPs each; logits averaged over both ensemble levels\\
\addlinespace

Fig.~\ref{fig:scaling_law}
& MNIST with resized feature dimensions $28^2$, $43^2$, $55^2$, and $64^2$, using bilinear interpolation; replacement noise at $p=0.97$; same class balancing, normalization and corruption conventions as Fig.~\ref{fig:actual_vs_predicted}
& Width-2048, 3-layer $(L=3)$ erf MLP; MSE loss; Adam; 25 epochs; batch size 100; learning rate $10^{-4}$; $N\in\{1000,2000,3000,4000\}$; 20 noisy datasets, 10 MLPs each per $(N,d)$. Accuracy uses initialization-averaged logits on all 10,000 clean test images; SNR uses the same 500 randomly sampled clean-test images across $(N,d)$ \\
\addlinespace

Fig.~\ref{fig:sbs_ptrain_ptest_auc}
& Same MNIST dataset, raw-input convention, and independent train/test replacement corruption as Fig.~\ref{fig:ptrain_ptest}
& Same trained models and 20 repeats per $p_{\rm train}$ as Fig.~\ref{fig:ptrain_ptest}; 21 training and 81 test corruption values. AUC computed by trapezoidal integration over $p_{\rm test}$ for each run; mean and standard error across runs \\
\addlinespace

Fig.~\ref{fig:actual_vs_predicted_additive_gaussian}
& MNIST, Fashion-MNIST, KMNIST; $N=4000$; $p=0.90,\ldots,1.00$ in steps of $0.01$; same splits and clean-image normalization as Fig.~\ref{fig:actual_vs_predicted}. Noisy training inputs use $(1-p)x+p\xi$, $\xi\sim\mathcal{N}(0,I)$, with no post-corruption renormalization; noise fixed across epochs; clean test sets
& Same width-2048, $L=3$ erf MLP, MSE loss, Adam settings, 30 epochs, batch size 250, and evaluation as Fig.~\ref{fig:actual_vs_predicted}. Accuracy: 20 noisy datasets, 10 MLPs each; inset and calibration at $p=0.97$: 100 noisy datasets, 50 MLPs each \\
\addlinespace

Fig.~\ref{fig:test_acc_vs_N_d_additive_gaussian}
& MNIST; same $(N,d)$ sweep and resizing as Fig.~\ref{fig:scaling_law}; additive-Gaussian noise at $p=0.97$; same normalization and corruption conventions as Fig.~\ref{fig:actual_vs_predicted_additive_gaussian}
& Same width-2048, $L=3$ erf MLP, MSE loss, Adam settings, 25 epochs, and batch size 100 as Fig.~\ref{fig:scaling_law}; 20 noisy datasets and 10 MLPs each per $(N,d)$. Same evaluation: 500 shared clean-test images for SNR; all 10,000 for accuracy using initialization-averaged logits \\
\addlinespace

Fig.~\ref{fig:centroid_vs_actual_depth=3_p=0-1}
& MNIST, Fashion-MNIST, KMNIST; $N=4000$, $d=784$; clean test sets. Same splits, normalization, and respective noise conventions as Figs.~\ref{fig:actual_vs_predicted} and~\ref{fig:actual_vs_predicted_additive_gaussian}. Both grids contain $0,0.1,\ldots,0.8$, followed by $0.90,0.91,\ldots,1.00$ for replacement and $0.85,0.86,\ldots,1.00$ for additive Gaussian
& Width-2048, $L=3$ erf MLP; MSE loss; Adam; 30 epochs; batch size 250; learning rate $10^{-4}$; 20 noisy datasets and 10 MLPs each per $p$. Evaluation matches the high-noise accuracy panels; their observations are reused. Effective model calibrated at $p=0.97$, not refitted at each $p$ \\
\addlinespace

Fig.~\ref{fig:centroid_vs_actual_depth=5}
& MNIST, Fashion-MNIST, KMNIST; replacement and additive-Gaussian noise at $p=0.90,0.91,\ldots,1.00$; $N=4000$, $d=784$; clean test sets. Same splits, normalization, and respective corruption conventions as Figs.~\ref{fig:actual_vs_predicted} and~\ref{fig:actual_vs_predicted_additive_gaussian}
& Width-2048, 5-layer $(L=5)$ ReLU MLP, counting the output layer; MSE loss; Adam; 30 epochs. Accuracy: 20 noisy datasets, 10 MLPs each; insets and architecture-specific calibration at $p=0.97$: 100 noisy datasets, 50 MLPs each. Remaining training and evaluation settings match the depth-$3$ experiments \\
\addlinespace

Fig.~\ref{fig:test_acc_activation}
& Mean-zero, unit-RMS normalized balanced subset of 4000 MNIST training examples; additive-Gaussian noise as in Fig.~\ref{fig:actual_vs_predicted_additive_gaussian}, fixed across epochs; 10,000 clean normalized test images
& Width-2048, 3-layer $(L=3)$ MLPs with erf, GELU, Swish, and Tanh; MSE loss; Adam; 20 epochs; batch size 250; learning rate $10^{-4}$; 50 noisy datasets with one initialization each per activation and $p=0.90,\ldots,1.00$ in steps of $0.01$; mean individual-model accuracy and sample SD \\

\bottomrule
\end{tabular}
\end{table*}

\newpage

\section{Further analysis of simultaneous train-time and test-time corruption}\label{app:ptrain_ptest}

\begin{figure*}[h]
    \centering
    \includegraphics[width=\linewidth]{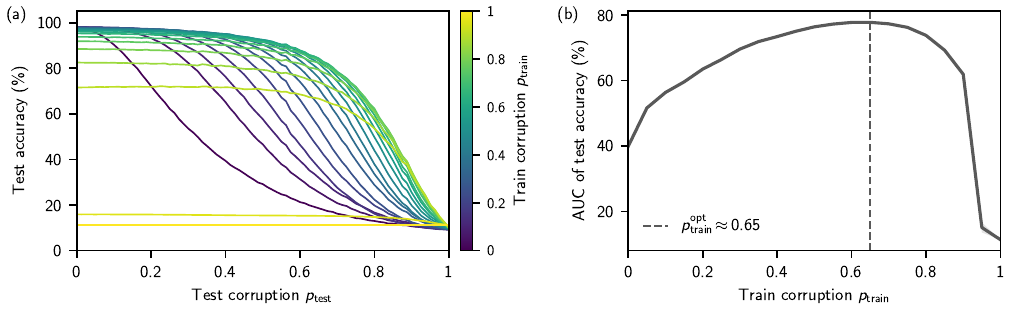}
    \caption{(a)~Test accuracy vs.\ test corruption $p_{\text{test}}$ for several values of the train corruption $p_{\text{train}}$. We find monotonically decreasing accuracy as a function of $p_{\text{test}}$, but for given $p_{\text{test}}$ the behavior with respect to $p_{\text{train}}$ is non-monotonic (see also Fig.~\ref{fig:ptrain_ptest} of the main text).
    (b)~Area under the curve (AUC) of the test accuracy vs.\ $p_{\text{test}}$, as a function of $p_{\text{train}}$. This is an overall measure of performance for a given $p_{\text{train}}$. We find that the AUC peaks around $p_{\text{train}}^{\text{opt }}\approx0.65$.
    All network parameters are the same as in Fig.~\ref{fig:ptrain_ptest} of the main text.
    }
    \label{fig:sbs_ptrain_ptest_auc}
\end{figure*}

In Sec.~\ref{sec:numerics} of the main text, we showed results for train-time corruption as well as test-time corruption, characterized by corruption levels $p_{\text{train}}$ and $p_{\text{test}}$, which are in general different.
We found that for a given value of $p_{\text{test}}$, the optimal training corruption level is $p_{\text{train}}\simeq p_{\text{test}}$.
Figure~\figref{fig:sbs_ptrain_ptest_auc}{a} shows the complementary view: accuracy as a function of $p_{\text{test}}$ for different values of $p_{\text{train}}$.

There, we find monotonic behavior: it is always best to test on clean data ($p_{\text{test}}=0$), indicating an asymmetry in the effect of train-time and test-time corruption.
Networks trained on corrupted data doing better when tested on clean data makes sense from the perspective of centroid classification: testing on actual examples from the clean distribution is optimal for the nearest-mean-class classifier.

As an overall metric of performance of a network with given $p_{\text{train}}$, we calculate the area under the curve (AUC) of the test accuracy vs.\ $p_{\text{test}}$.
Figure~\figref{fig:sbs_ptrain_ptest_auc}{b} shows the AUC as a function of $p_{\text{train}}$, exhibiting non-monotonic behavior with a peak around $p_{\text{train}}^{\text{opt }}\approx0.65$.
This implies that, if the test-time corruption is not known ahead of time, training with corruption level $\approx0.65$ will likely lead to optimal performance.


\section{High-noise NTK expansion}\label{app:NTK}

\allowdisplaybreaks

\subsection{Setup}

We consider a training set with $N$ examples, $C$ classes, and feature dimension $d$.
The clean training features are denoted by $\mathbf{x}\in\mathbb{R}^{d\times N}$, while the corrupted training features are denoted by $\tilde{\mathbf{x}}\in\mathbb{R}^{d\times N}$.
The clean labels are denoted by $\mathbf{y}\in\mathbb{R}^{N\times C}$, with $\boldsymbol{y}_i\in\mathbb{R}^N$ the one-hot target vector for class $i=1,\ldots,C$.
A clean test feature is denoted by $x_*\in\mathbb{R}^d$. The notation in this section is self-contained.

The index $\chi$ is reserved for training examples.
Other Greek indices, such as $\alpha,\beta,\gamma,\delta$, may refer either to training examples or to the test point $*$.
We write $z_\alpha$ for the input associated with index $\alpha$, so that $z_\chi=\tilde{x}_\chi$ for a corrupted training input and $z_*=x_*$ for the clean test input.

We work in the infinite-width NTK limit for an arbitrary $L$-layer MLP with activation function $\sigma(\cdot)$.
The weights and biases at initialization are drawn i.i.d.\ from mean-zero Gaussian distributions with variances $C_w$ and $C_b$, respectively.
In this limit, training with squared loss gives the class-$i$ output
\begin{equation}
    f_i(x_*)=
    \tilde{\boldsymbol{\Theta}}_{*\chi}^{(L)}
    \left[\tilde{\boldsymbol{\Theta}}_{\chi\chi}^{(L)}\right]^{-1}
    \boldsymbol{y}_i,
    \label{eq:app_output_general}
\end{equation}
where $\tilde{\boldsymbol{\Theta}}_{\chi\chi}^{(L)}\in\mathbb{R}^{N\times N}$ is the training--training NTK and $\tilde{\boldsymbol{\Theta}}_{*\chi}^{(L)}\in\mathbb{R}^{1\times N}$ is the test-training NTK.
The tilde indicates that the NTK is evaluated on the corrupted training features.

For arbitrary functions $g$ and $h$, we define the two-dimensional Gaussian expectation by
\begin{equation}
    \langle g_\alpha h_\beta\rangle_K
    =
    \frac{1}{2\pi\sqrt{\det K}}
    \int da\,db\,
    g(a)h(b)
    \exp\left[
        -\frac{1}{2}
        \begin{pmatrix}
            a & b
        \end{pmatrix}
            \begin{pmatrix}
        K_{\alpha\alpha} & K_{\alpha\beta} \\
        K_{\beta\alpha} & K_{\beta\beta}
    \end{pmatrix}^{-1}
        \begin{pmatrix}
            a \\ b
        \end{pmatrix}
    \right].
    \label{eq:app_2d_gaussian_expectation}
\end{equation}

The corresponding one-dimensional Gaussian expectation is denoted by
\begin{equation}
    \langle gh\rangle_{K_{\alpha\alpha}}
    =
    \frac{1}{\sqrt{2\pi K_{\alpha\alpha}}}
    \int da\,g(a)h(a)
    \exp\left[-\frac{a^2}{2K_{\alpha\alpha}}\right].
    \label{eq:app_1d_gaussian_expectation}
\end{equation}

The noisy Bayesian kernel is defined recursively as
\begin{equation}
    \tilde{K}_{\alpha\beta}^{(l+1)}
    =
    C_b+C_w\langle \sigma_\alpha\sigma_\beta\rangle_{\tilde{K}^{(l)}},
    \label{eq:app_bayesian_kernel_recursion}
\end{equation}
with initial condition
\begin{equation}
    \tilde{K}_{\alpha\beta}^{(1)}
    =
    C_b+\frac{C_w}{d}\sum_{j=1}^d z_{j\alpha}z_{j\beta}.
    \label{eq:app_bayesian_kernel_initial}
\end{equation}
The noisy NTK is defined recursively by
\begin{equation}
    \tilde{\Theta}_{\alpha\beta}^{(l+1)}
    =
    C_b+C_w\left(
        \langle \sigma_\alpha\sigma_\beta\rangle_{\tilde{K}^{(l)}}
        +
        \langle \sigma'_\alpha\sigma'_\beta\rangle_{\tilde{K}^{(l)}}
        \tilde{\Theta}_{\alpha\beta}^{(l)}
    \right),
    \label{eq:app_ntk_recursion}
\end{equation}
with initial condition
\begin{equation}
    \tilde{\Theta}_{\alpha\beta}^{(1)}
    =
    \tilde{K}_{\alpha\beta}^{(1)}.
    \label{eq:app_ntk_initial}
\end{equation}

Since we work in the large-$d$ limit, the initial feature average obeys a central-limit expansion,
\begin{equation}
    \frac{C_w}{d}\sum_{j=1}^{d}z_{j\alpha} z_{j\beta}
    =
    \frac{C_w}{d}\sum_{j=1}^d \mathbb{E}[z_{j\alpha}z_{j\beta}]
    +
    \frac{1}{\sqrt d}\eta_{\alpha\beta}
    +
    \mathcal{O}\left(d^{-1}\right),
    \label{eq:app_initial_clt}
\end{equation}
where $\eta_{\alpha\beta}$ is a mean-zero Gaussian random tensor with covariance determined by the noise model.
Here and throughout this appendix, $\mathbb{E}[\cdot]$ denotes expectation over the input-corruption noise.
Thus the first-layer Bayesian kernel may be decomposed as
\begin{equation}
    \tilde{K}_{\alpha\beta}^{(1)}
    =
    K_{\alpha\beta}^{(1)}
    +
    \frac{1}{\sqrt d}\eta_{\alpha\beta}
    +
    \mathcal{O}\left(d^{-1}\right),
    \label{eq:app_first_layer_decomposition}
\end{equation}
where
\begin{equation}
    K_{\alpha\beta}^{(1)}
    =
    C_b+\frac{C_w}{d}\sum_{j=1}^d \mathbb{E}[z_{j\alpha}z_{j\beta}].
    \label{eq:app_first_layer_mean_kernel}
\end{equation}

The corruption model is assumed to have a noise strength $p\in[0,1]$.
We define the signal strength
\begin{equation}
    \epsilon=1-p,
    \label{eq:app_epsilon_definition}
\end{equation}
so that $\epsilon=0$ is the pure-noise limit.
Both $K_{\alpha\beta}^{(1)}$ and $\eta_{\alpha\beta}$ may depend on $\epsilon$.
Since the expansion is around the pure-noise point, all quantities below are evaluated at $\epsilon=0$ unless otherwise specified.
For example,
\begin{equation}
    K_{\alpha\beta}^{(1)}
    =
    K_{\alpha\beta}^{(1)}\big|_{\epsilon=0},
    \qquad
    \frac{\partial K_{\alpha\beta}^{(1)}}{\partial\epsilon}
    =
    \left.
    \frac{\partial K_{\alpha\beta}^{(1)}(\epsilon)}{\partial\epsilon}
    \right|_{\epsilon=0}.
    \label{eq:app_pure_noise_convention}
\end{equation}
With this convention, the first-layer perturbation has the form
\begin{equation}
    \delta K_{\alpha\beta}^{(1)}
    =
    \epsilon\frac{\partial K_{\alpha\beta}^{(1)}}{\partial\epsilon}
    +
    \frac{1}{\sqrt d}\eta_{\alpha\beta}.
    \label{eq:app_first_layer_perturbation}
\end{equation}

\subsection{Useful identities}

We collect several identities used repeatedly below.
First, Price's theorem gives the following derivatives of Gaussian expectations with respect to the covariance.
For one-dimensional Gaussian expectations,
\begin{align}
\frac{\partial}{\partial K_{\alpha\alpha}}
\langle g h \rangle_{K_{\alpha\alpha}}
&=
\langle g' h' \rangle_{K_{\alpha\alpha}}
+
\frac{1}{2}\langle g'' h \rangle_{K_{\alpha\alpha}}
+
\frac{1}{2}\langle g h'' \rangle_{K_{\alpha\alpha}}.
\label{eq:app_price_1d}
\end{align}
For two-dimensional Gaussian expectations with $\alpha\neq\beta$,
\begin{align}
\frac{\partial}{\partial K_{\alpha\alpha}}
\langle g_\alpha h_\beta \rangle_K
&=
\frac{1}{2}\langle g''_\alpha h_\beta \rangle_K,
\\
\frac{\partial}{\partial K_{\alpha\beta}}
\langle g_\alpha h_\beta \rangle_K
&=
\langle g'_\alpha h'_\beta \rangle_K,
\\
\frac{\partial}{\partial K_{\beta\beta}}
\langle g_\alpha h_\beta \rangle_K
&=
\frac{1}{2}\langle g_\alpha h''_\beta \rangle_K.
\label{eq:app_price_2d}
\end{align}
Although these expressions are written for smooth activations, non-smooth activations such as ReLU may be recovered by a limiting procedure.

We also use the elementary solution of the scalar recursion
\begin{equation}
    B^{(l+1)}=a_l+b_lB^{(l)}.
\end{equation}
It is
\begin{equation}
    B^{(l)}
    =
    \sum_{i=1}^{l-1}a_i\prod_{j=i+1}^{l-1}b_j
    +
    B^{(1)}\prod_{i=1}^{l-1}b_i.
    \label{eq:app_scalar_recursion_solution}
\end{equation}

Next, for an $N\times N$ matrix of the form
\begin{equation}
    H=h_d\mathbf{I}+h_o(\mathbf{1}\mathbf{1}^{\mathrm{T}}-\mathbf{I}),
\end{equation}
where $\mathbf{I}$ is the identity and $\mathbf{1}$ is the vector of all ones, the inverse is
\begin{equation}
    H^{-1}
    =
    \frac{1}{h_d-h_o}
    \left(
        \mathbf{I}
        -
        \frac{h_o}{h_d-h_o+h_oN}
        \mathbf{1}\mathbf{1}^{\mathrm{T}}
    \right).
    \label{eq:app_constant_matrix_inverse}
\end{equation}
For vectors $\mathbf{a},\mathbf{b}\in\mathbb{R}^N$, defining
\begin{equation}
    \mathrm{E}_\alpha[a_\alpha]
    =
    \frac{1}{N}\sum_{\alpha=1}^N a_\alpha,
    \qquad
    \mathrm{E}_\alpha[a_\alpha b_\alpha]
    =
    \frac{1}{N}\sum_{\alpha=1}^N a_\alpha b_\alpha,
\end{equation}
we therefore have
\begin{equation}
    \mathbf{a}^{\mathrm{T}}H^{-1}\mathbf{b}
    =
    \frac{N}{h_d-h_o}
    \left(
        \mathrm{E}_\alpha[a_\alpha b_\alpha]
        -
        \frac{h_oN}{h_d-h_o+h_oN}
        \mathrm{E}_\alpha[a_\alpha]\mathrm{E}_\beta[b_\beta]
    \right).
    \label{eq:app_inverse_contraction_identity}
\end{equation}

The following layer-to-layer quantities will also be used:
\begin{equation}
    P_{\alpha\beta}^{a\rightarrow b}
    =
    \prod_{i=a}^{b-1}
    \langle \sigma'_\alpha\sigma'_\beta\rangle_{K^{(i)}},
    \label{eq:app_P_definition}
\end{equation}
\begin{equation}
    U_{\alpha}^{a\rightarrow b}
    =
    \prod_{i=a}^{b-1}
    \left(
        \langle \sigma'\sigma'\rangle_{K_{\alpha\alpha}^{(i)}}
        +
        \langle \sigma\sigma''\rangle_{K_{\alpha\alpha}^{(i)}}
    \right),
    \label{eq:app_U_definition}
\end{equation}
and
\begin{equation}
    M_{\alpha\beta}^{a\rightarrow b}
    =
    \frac{1}{2}
    \sum_{i=a}^{b-1}
    \langle \sigma''_\alpha\sigma_\beta\rangle_{K^{(i)}}
    U_\alpha^{a\rightarrow i}
    P_{\alpha\beta}^{i+1\rightarrow b}.
    \label{eq:app_M_definition}
\end{equation}
The corresponding NTK-response coefficients are
\begin{align}
R_\alpha^{a\rightarrow b}
&=
\sum_{i=a}^{b-1}
\begin{aligned}[t]
\Big(
&\langle \sigma'\sigma'\rangle_{K_{\alpha\alpha}^{(i)}}
+
\langle \sigma\sigma''\rangle_{K_{\alpha\alpha}^{(i)}}
\\
&+
\left[
    \langle \sigma''\sigma''\rangle_{K_{\alpha\alpha}^{(i)}}
    +
    \langle \sigma'\sigma'''\rangle_{K_{\alpha\alpha}^{(i)}}
\right]
\Theta_{\alpha\alpha}^{(i)}
\Big)
U_\alpha^{a\rightarrow i}
P_{\alpha\alpha}^{i+1\rightarrow b} +
P_{\alpha\alpha}^{a\rightarrow b},
\end{aligned}
\\
S_{\alpha\beta}^{a\rightarrow b}
&=
\sum_{i=a}^{b-1}
\begin{aligned}[t]
\Bigg[
&\frac{1}{2}
\left(
    \langle \sigma''_\alpha\sigma_\beta\rangle_{K^{(i)}}
    +
    \Theta_{\alpha\beta}^{(i)}
    \langle \sigma'''_\alpha\sigma'_\beta\rangle_{K^{(i)}}
\right)
U_\alpha^{a\rightarrow i}
\\
&+
\left(
    \langle \sigma'_\alpha\sigma'_\beta\rangle_{K^{(i)}}
    +
    \Theta_{\alpha\beta}^{(i)}
    \langle \sigma''_\alpha\sigma''_\beta\rangle_{K^{(i)}}
\right)
M_{\alpha\beta}^{a\rightarrow i}
\Bigg]
P_{\alpha\beta}^{i+1\rightarrow b},
\end{aligned}
\\
T_{\alpha\beta}^{a\rightarrow b}
&=
\sum_{i=a}^{b-1}
\left(
    \langle \sigma'_\alpha\sigma'_\beta\rangle_{K^{(i)}}
    +
    \Theta_{\alpha\beta}^{(i)}
    \langle \sigma''_\alpha\sigma''_\beta\rangle_{K^{(i)}}
\right)
P_{\alpha\beta}^{a\rightarrow i}
P_{\alpha\beta}^{i+1\rightarrow b}
+
P_{\alpha\beta}^{a\rightarrow b}.
\label{eq:app_RST_definitions}
\end{align}
All quantities in Eqs.~\eqref{eq:app_P_definition}--\eqref{eq:app_RST_definitions} are evaluated in the pure-noise limit.
Therefore their values depend only on whether the indices are training--training, test-training, or test-test indices; they do not otherwise depend on the individual data points.

\subsection{Influence of input perturbations on the output}

We now compute how a perturbation in the first-layer Bayesian kernel propagates to the output.
Let
\begin{equation}
    \tilde{K}_{\alpha\beta}^{(l)}
    =
    K_{\alpha\beta}^{(l)}
    +
    \delta K_{\alpha\beta}^{(l)}
    +
    \mathcal{O}(\delta K^2),
    \qquad
    \tilde{\Theta}_{\alpha\beta}^{(l)}
    =
    \Theta_{\alpha\beta}^{(l)}
    +
    \delta \Theta_{\alpha\beta}^{(l)}
    +
    \mathcal{O}(\delta K^2).
    \label{eq:app_kernel_perturbation_expansion}
\end{equation}
The unperturbed kernels satisfy the pure-noise recursions
\begin{align}
\Theta_{\alpha\beta}^{(l+1)}
&=
C_b+C_w
\left(
    \langle \sigma_\alpha\sigma_\beta\rangle_{K^{(l)}}
    +
    \langle \sigma'_\alpha\sigma'_\beta\rangle_{K^{(l)}}
    \Theta_{\alpha\beta}^{(l)}
\right),
\\
K_{\alpha\beta}^{(l+1)}
&=
C_b+C_w
\langle \sigma_\alpha\sigma_\beta\rangle_{K^{(l)}}.
\label{eq:app_unperturbed_recursions}
\end{align}
The linear perturbations obey
\begin{align}
\frac{1}{C_w}\delta K^{(l+1)}_{\alpha\beta}
&=
\delta_{\alpha\beta}
\left(
    \langle \sigma'\sigma'\rangle_{K_{\alpha\alpha}^{(l)}}
    +
    \langle \sigma\sigma''\rangle_{K_{\alpha\alpha}^{(l)}}
\right)
\delta K_{\alpha\alpha}^{(l)}
\notag \\
&\quad+
(1-\delta_{\alpha\beta})
\Bigg[
    \frac{1}{2}
    \langle \sigma''_\alpha\sigma_\beta\rangle_{K^{(l)}}
    \delta K_{\alpha\alpha}^{(l)}
    +
    \langle \sigma'_\alpha\sigma'_\beta\rangle_{K^{(l)}}
    \delta K_{\alpha\beta}^{(l)}
    +
    \frac{1}{2}
    \langle \sigma_\alpha\sigma''_\beta\rangle_{K^{(l)}}
    \delta K_{\beta\beta}^{(l)}
\Bigg],
\\
\frac{1}{C_w}\delta \Theta^{(l+1)}_{\alpha\beta}
&=
\delta_{\alpha\beta}
\Bigg[
\Big(
    \langle \sigma'\sigma'\rangle_{K_{\alpha\alpha}^{(l)}}
    +
    \langle \sigma\sigma''\rangle_{K_{\alpha\alpha}^{(l)}}
\notag \\
&\qquad\qquad\qquad
    +
    \left[
        \langle \sigma''\sigma''\rangle_{K_{\alpha\alpha}^{(l)}}
        +
        \langle \sigma'\sigma'''\rangle_{K_{\alpha\alpha}^{(l)}}
    \right]
    \Theta_{\alpha\alpha}^{(l)}
\Big)
\delta K_{\alpha\alpha}^{(l)}
+
\langle \sigma'\sigma'\rangle_{K_{\alpha\alpha}^{(l)}}
\delta \Theta_{\alpha\alpha}^{(l)}
\Bigg]
\notag \\
&\quad+
(1-\delta_{\alpha\beta})
\Bigg[
    \frac{1}{2}
    \left(
        \langle \sigma''_\alpha\sigma_\beta\rangle_{K^{(l)}}
        +
        \langle \sigma'''_\alpha\sigma'_\beta\rangle_{K^{(l)}}
        \Theta_{\alpha\beta}^{(l)}
    \right)
    \delta K_{\alpha\alpha}^{(l)}
\notag \\
&\qquad\qquad
    +
    \left(
        \langle \sigma'_\alpha\sigma'_\beta\rangle_{K^{(l)}}
        +
        \langle \sigma''_\alpha\sigma''_\beta\rangle_{K^{(l)}}
        \Theta_{\alpha\beta}^{(l)}
    \right)
    \delta K_{\alpha\beta}^{(l)}
\notag \\
&\qquad\qquad
    +
    \frac{1}{2}
    \left(
        \langle \sigma_\alpha\sigma''_\beta\rangle_{K^{(l)}}
        +
        \langle \sigma'_\alpha\sigma'''_\beta\rangle_{K^{(l)}}
        \Theta_{\alpha\beta}^{(l)}
    \right)
    \delta K_{\beta\beta}^{(l)}
\notag \\
&\qquad\qquad
    +
    \langle \sigma'_\alpha\sigma'_\beta\rangle_{K^{(l)}}
    \delta \Theta_{\alpha\beta}^{(l)}
\Bigg].
\label{eq:app_linearized_recursions}
\end{align}

Given the solution of the pure-noise Bayesian-kernel recursion, the pure-noise NTK has the closed-form solution
\begin{equation}
    \frac{1}{C_w^{l-1}}\Theta_{\alpha\beta}^{(l)}
    =
    \sum_{i=1}^{l-1}
    \left(
        C_b+C_w\langle \sigma_\alpha\sigma_\beta\rangle_{K^{(i)}}
    \right)
    C_w^{-i}
    P_{\alpha\beta}^{i+1\rightarrow l}
    +
    P_{\alpha\beta}^{1\rightarrow l}
    \Theta_{\alpha\beta}^{(1)},
    \label{eq:app_ntk_solution}
\end{equation}
and the perturbations propagate as
\begin{align}
\frac{1}{C_w^{l-1}}\delta K_{\alpha\beta}^{(l)}
&=
\delta_{\alpha\beta}
U_\alpha^{1\rightarrow l}
\delta K_{\alpha\alpha}^{(1)}
\notag \\
&\quad+
(1-\delta_{\alpha\beta})
\left(
    M_{\alpha\beta}^{1\rightarrow l}
    \delta K_{\alpha\alpha}^{(1)}
    +
    M_{\beta\alpha}^{1\rightarrow l}
    \delta K_{\beta\beta}^{(1)}
    +
    P_{\alpha\beta}^{1\rightarrow l}
    \delta K_{\alpha\beta}^{(1)}
\right),
\\
\frac{1}{C_w^{l-1}}\delta\Theta_{\alpha\beta}^{(l)}
&=
\delta_{\alpha\beta}
R_\alpha^{1\rightarrow l}
\delta K_{\alpha\alpha}^{(1)}
\notag \\
&\quad+
(1-\delta_{\alpha\beta})
\left(
    S_{\alpha\beta}^{1\rightarrow l}
    \delta K_{\alpha\alpha}^{(1)}
    +
    S_{\beta\alpha}^{1\rightarrow l}
    \delta K_{\beta\beta}^{(1)}
    +
    T_{\alpha\beta}^{1\rightarrow l}
    \delta K_{\alpha\beta}^{(1)}
\right).
\label{eq:app_delta_kernel_solutions}
\end{align}

At pure noise, the training--training NTK at layer $L$ has the permutation-symmetric form
\begin{equation}
    \frac{1}{C_w^{L-1}}\boldsymbol{\Theta}_{\chi\chi}^{(L)}
    =
    \theta_d\mathbf{I}
    +
    \theta_o(\mathbf{1}\mathbf{1}^{\mathrm{T}}-\mathbf{I}),
    \label{eq:app_pure_noise_train_ntk}
\end{equation}
while the test-training NTK has the form
\begin{equation}
    \frac{1}{C_w^{L-1}}\boldsymbol{\Theta}_{*\chi}^{(L)}
    =
    \theta_*\mathbf{1}^{\mathrm{T}}.
    \label{eq:app_pure_noise_test_ntk}
\end{equation}
Here $\theta_d$, $\theta_o$, and $\theta_*$ are constants determined by the architecture and the pure-noise kernel.
We define
\begin{equation}
    \psi
    =
    \frac{N}{\theta_d-\theta_o+\theta_o N},
    \qquad
    \mathrm{Cov}_\alpha^1(a_\alpha,b_\alpha)
    =
    \mathrm{E}_\alpha[a_\alpha b_\alpha]
    -
    \theta_o\psi\,
    \mathrm{E}_\alpha[a_\alpha]\mathrm{E}_\alpha[b_\alpha].
    \label{eq:app_psi_cov1_definition}
\end{equation}

Expanding the kernel-regression output to first order in $\delta\Theta$ gives
\begin{equation}
    f_i(x_*)
    =
    \left[
        \boldsymbol{\Theta}_{*\chi}^{(L)}
        \left[\boldsymbol{\Theta}_{\chi\chi}^{(L)}\right]^{-1}
        +
        \delta\boldsymbol{\Theta}_{*\chi}^{(L)}
        \left[\boldsymbol{\Theta}_{\chi\chi}^{(L)}\right]^{-1}
        -
        \boldsymbol{\Theta}_{*\chi}^{(L)}
        \left[\boldsymbol{\Theta}_{\chi\chi}^{(L)}\right]^{-1}
        \delta\boldsymbol{\Theta}_{\chi\chi}^{(L)}
        \left[\boldsymbol{\Theta}_{\chi\chi}^{(L)}\right]^{-1}
    \right]\boldsymbol{y}_i.
    \label{eq:app_output_first_order_delta}
\end{equation}
Substituting the pure-noise inverse and the perturbative solution gives
\begin{align}
f_i(x_*)
=&\
\theta_*\psi\,\mathrm{E}_\alpha[y_{\alpha i}]
\notag \\
&+
\frac{N}{\theta_d-\theta_o}
\Bigg(
    \left(
        S_{\chi *}^{1\rightarrow L}
        -
        \theta_*\psi S_{\chi\chi}^{1\rightarrow L}
    \right)
    \mathrm{Cov}_\alpha^1
    \left(
        \delta K_{\alpha\alpha}^{(1)},
        y_{\alpha i}
    \right)
\notag \\
&\qquad\qquad\qquad
    +
    T_{*\chi}^{1\rightarrow L}
    \mathrm{Cov}_\alpha^1
    \left(
        \delta K_{*\alpha}^{(1)},
        y_{\alpha i}
    \right)
\notag \\
&\qquad\qquad\qquad
    -
    \theta_*\psi
    \Bigg[
        T_{\chi\chi}^{1\rightarrow L}
        \mathrm{E}_\alpha
        \left[
            \mathrm{Cov}_\beta^1
            \left(
                \delta K_{\alpha\beta}^{(1)},
                y_{\beta i}
            \right)
        \right]
\notag \\
&\qquad\qquad\qquad\qquad\quad
        +
        S_{\chi\chi}^{1\rightarrow L}
        \left(1-\theta_o\psi\right)
        \mathrm{E}_\alpha
        \left[
            \delta K_{\alpha\alpha}^{(1)}
        \right]
        \mathrm{E}_\alpha[y_{\alpha i}]
    \Bigg]
\Bigg)
+
\mathcal{O}(\delta K\,N^0)
+
\mathcal{O}(\delta K^2).
\label{eq:app_general_delta_output}
\end{align}

\subsection{Output fluctuations under Gaussian kernel perturbations}

We now isolate the fluctuation term obtained by taking
\begin{equation}
    \delta K_{\alpha\beta}^{(1)}
    =
    \frac{1}{\sqrt d}\eta_{\alpha\beta},
    \label{eq:app_eta_only_perturbation}
\end{equation}
where $\eta_{\alpha\beta}$ is a mean-zero Gaussian random tensor.
Since Eq.~\eqref{eq:app_general_delta_output} is linear in $\eta_{\alpha\beta}$, the induced output fluctuation is also Gaussian.
It is therefore enough to compute its covariance.

The basic identity is
\begin{equation}
    \mathbb{E}
    \left[
        \mathrm{Cov}_\beta^1(\eta_{\alpha\beta},y_{\beta i})
        \mathrm{Cov}_\delta^1(\eta_{\gamma\delta},y_{\delta j})
    \right]
    =
    \frac{1}{N}
    \mathbb{C}\mathrm{ov}(\eta_{\alpha\beta},\eta_{\beta\gamma})
    \mathrm{Cov}_\alpha^2(y_{\alpha i},y_{\alpha j}),
    \label{eq:app_basic_eta_identity}
\end{equation}
where
\begin{equation}
    \mathrm{Cov}_\alpha^2(a_\alpha,b_\alpha)
    =
    \mathrm{E}_\alpha[a_\alpha b_\alpha]
    -
    \left(
        2\theta_o\psi-(\theta_o\psi)^2
    \right)
    \mathrm{E}_\alpha[a_\alpha]\mathrm{E}_\alpha[b_\alpha].
    \label{eq:app_cov2_definition}
\end{equation}
This identity follows from the i.i.d.\ structure of the input noise: a covariance between two $\eta$ variables vanishes unless the two index pairs share at least one data index.

For large $N$, the required contractions are
\begin{align}
\mathbb{E}
\left[
    \mathrm{Cov}_\alpha^1(\eta_{\alpha\alpha},y_{\alpha i})
    \mathrm{Cov}_\beta^1(\eta_{\beta\beta},y_{\beta j})
\right]
&=
\frac{1}{N}
\mathbb{V}\mathrm{ar}(\eta_{\alpha\alpha})
\mathrm{Cov}_\beta^2(y_{\beta i},y_{\beta j}),
\\
\mathbb{E}
\left[
    \mathrm{Cov}_\alpha^1(\eta_{\alpha\alpha},y_{\alpha i})
    \mathrm{Cov}_\beta^1(\eta_{*\beta},y_{\beta j})
\right]
&=
\frac{1}{N}
\mathbb{C}\mathrm{ov}(\eta_{\alpha\alpha},\eta_{*\alpha})
\mathrm{Cov}_\beta^2(y_{\beta i},y_{\beta j}),
\\
\mathbb{E}
\left[
    \mathrm{Cov}_\alpha^1(\eta_{\alpha\alpha},y_{\alpha i})
    \mathrm{E}_\beta
    \left[
        \mathrm{Cov}_\gamma^1(\eta_{\beta\gamma},y_{\gamma j})
    \right]
\right]
&=
\frac{1}{N}
\mathbb{C}\mathrm{ov}(\eta_{\alpha\alpha},\eta_{\alpha\beta})
\notag \\
&\quad\times
\Bigg(
    \mathrm{Cov}_\gamma^2(y_{\gamma i},y_{\gamma j})
    +
    (1-\theta_o\psi)^2
    \mathrm{E}_\gamma[y_{\gamma i}]
    \mathrm{E}_\delta[y_{\delta j}]
\Bigg)
\notag \\
&\quad+
\mathcal{O}\left(\frac{1}{N^2}\right),
\\
\mathbb{E}
\left[
    \mathrm{Cov}_\alpha^1(\eta_{\alpha\alpha},y_{\alpha i})
    \mathrm{E}_\beta[\eta_{\beta\beta}]
    \mathrm{E}_\gamma[y_{\gamma j}]
\right]
&=
\frac{1}{N}
\mathbb{V}\mathrm{ar}(\eta_{\alpha\alpha})
(1-\theta_o\psi)
\mathrm{E}_\beta[y_{\beta i}]
\mathrm{E}_\gamma[y_{\gamma j}],
\\
\mathbb{E}
\left[
    \mathrm{Cov}_\alpha^1(\eta_{*\alpha},y_{\alpha i})
    \mathrm{Cov}_\beta^1(\eta_{*\beta},y_{\beta j})
\right]
&=
\frac{1}{N}
\mathbb{V}\mathrm{ar}(\eta_{*\alpha})
\mathrm{Cov}_\beta^2(y_{\beta i},y_{\beta j}),
\\
\mathbb{E}
\left[
    \mathrm{Cov}_\alpha^1(\eta_{*\alpha},y_{\alpha i})
    \mathrm{E}_\beta
    \left[
        \mathrm{Cov}_\gamma^1(\eta_{\beta\gamma},y_{\gamma j})
    \right]
\right]
&=
\frac{1}{N}
\mathbb{C}\mathrm{ov}(\eta_{*\alpha},\eta_{\alpha\beta})
\notag \\
&\quad\times
\Bigg(
    \mathrm{Cov}_\gamma^2(y_{\gamma i},y_{\gamma j})
    +
    (1-\theta_o\psi)^2
    \mathrm{E}_\gamma[y_{\gamma i}]
    \mathrm{E}_\delta[y_{\delta j}]
\Bigg)
\notag \\
&\quad+
\mathcal{O}\left(\frac{1}{N^2}\right),
\\
\mathbb{E}
\left[
    \mathrm{Cov}_\alpha^1(\eta_{*\alpha},y_{\alpha i})
    \mathrm{E}_\beta[\eta_{\beta\beta}]
    \mathrm{E}_\gamma[y_{\gamma j}]
\right]
&=
\frac{1}{N}
\mathbb{C}\mathrm{ov}(\eta_{*\alpha},\eta_{\alpha\alpha})
(1-\theta_o\psi)
\mathrm{E}_\beta[y_{\beta i}]
\mathrm{E}_\gamma[y_{\gamma j}],
\\
\mathbb{E}
\left[
    \mathrm{E}_\alpha
    \left[
        \mathrm{Cov}_\beta^1(\eta_{\alpha\beta},y_{\beta i})
    \right]
    \mathrm{E}_\gamma
    \left[
        \mathrm{Cov}_\delta^1(\eta_{\gamma\delta},y_{\delta j})
    \right]
\right]
&=
\frac{1}{N}
\mathbb{C}\mathrm{ov}(\eta_{\alpha\beta},\eta_{\beta\gamma})
\notag \\
&\quad\times
\Bigg(
    \mathrm{Cov}_\delta^2(y_{\delta i},y_{\delta j})
    +
    3(1-\theta_o\psi)^2
    \mathrm{E}_\delta[y_{\delta i}]
    \mathrm{E}_\lambda[y_{\lambda j}]
\Bigg)
\notag \\
&\quad+
\mathcal{O}\left(\frac{1}{N^2}\right),
\\
\mathbb{E}
\left[
    \mathrm{E}_\alpha
    \left[
        \mathrm{Cov}_\beta^1(\eta_{\alpha\beta},y_{\beta i})
    \right]
    \mathrm{E}_\gamma[\eta_{\gamma\gamma}]
    \mathrm{E}_\delta[y_{\delta j}]
\right]
&=
\frac{2}{N}
\mathbb{C}\mathrm{ov}(\eta_{\alpha\beta},\eta_{\beta\beta})
(1-\theta_o\psi)
\mathrm{E}_\gamma[y_{\gamma i}]
\mathrm{E}_\delta[y_{\delta j}]
\notag \\
&\quad+
\mathcal{O}\left(\frac{1}{N^2}\right),
\\
\mathbb{E}
\left[
    \mathrm{E}_\alpha[\eta_{\alpha\alpha}]
    \mathrm{E}_\beta[y_{\beta i}]
    \mathrm{E}_\gamma[\eta_{\gamma\gamma}]
    \mathrm{E}_\delta[y_{\delta j}]
\right]
&=
\frac{1}{N}
\mathbb{V}\mathrm{ar}(\eta_{\alpha\alpha})
\mathrm{E}_\beta[y_{\beta i}]
\mathrm{E}_\gamma[y_{\gamma j}].
\label{eq:app_eta_contractions}
\end{align}
The variances and covariances appearing here are scalars evaluated at pure noise.
In particular, they have no explicit dependence on the clean training data, because the clean signal has been set to zero before evaluating the fluctuation covariance.

Combining these contractions gives
\begin{equation}
    \mathrm{Cov}\left(f_i(x_*),f_j(x_*)\right)
    =
    \frac{N/d}{(\theta_d-\theta_o)^2}
    \tilde{V}_{ij}
    +
    \mathcal{O}\left(\frac{1}{d^2}\right)
    +
    \mathcal{O}\left(\frac{N^0}{d}\right)
    +
    \mathcal{O}\left(\frac{N}{d^2}\right),
    \label{eq:app_output_covariance_general}
\end{equation}
where $\tilde{V}_{ij}$ is an architecture- and noise-dependent coefficient matrix. For C equally partitioned classes,
\begin{equation}
    \mathrm{Cov}_\alpha^2(y_{\alpha i},y_{\alpha j})
    =
    \frac{1}{C}
    \left(
        \delta_{ij}
        -
        \frac{1}{C}
        \left[
            2\theta_o\psi-(\theta_o\psi)^2
        \right]
    \right).
    \label{eq:app_class_covariance}
\end{equation}
Thus the coefficient matrix can be written as
\begin{equation}
    \tilde{V}_{ij}
    =
    \delta_{ij}\tilde{V}_d
    +
    (1-\delta_{ij})\tilde{V}_o,
    \label{eq:app_V_decomposition}
\end{equation}
for scalars $\tilde{V}_d$ and $\tilde{V}_o$. Since the $\mathrm{Cov}^2$ has an overall factor of $1/C$, we can pull this out by defining $V_{ij}=C\tilde{V}_{ij}$. Thus, the output fluctuations may be written as
\begin{equation}
    f_i(x_*)-\mathbb{E}[f_i(x_*)]
    =
    \sqrt{\frac{N/C}{d}}
    \frac{\sqrt{V_d}}{\theta_d-\theta_o}
    \hat{\xi}_i,
    \qquad
    \hat{\boldsymbol{\xi}}\sim\mathcal{N}(0,\Sigma),
    \label{eq:app_output_fluctuation_reparam}
\end{equation}
where
\begin{equation}
    \Sigma_{ij}
    =
    \delta_{ij}
    +
    (1-\delta_{ij})\frac{V_o}{V_d}.
    \label{eq:app_fluctuation_covariance_matrix}
\end{equation}

\subsection{The output for an arbitrary i.i.d.\ noise model}

For an arbitrary i.i.d.\ noise model, the first-layer perturbation is
\begin{equation}
    \delta K_{\alpha\beta}^{(1)}
    =
    \epsilon\frac{\partial K_{\alpha\beta}^{(1)}}{\partial\epsilon}
    +
    \frac{1}{\sqrt d}\eta_{\alpha\beta}.
    \label{eq:app_arbitrary_noise_first_layer}
\end{equation}
Substituting this into Eq.~\eqref{eq:app_general_delta_output}, and using the fluctuation result Eq.~\eqref{eq:app_output_fluctuation_reparam}, gives the general high-noise output. Here, we assume the dataset has been partitioned into $C$ equally represented classes. We define
\begin{equation}
    \Delta\mathrm{E}_\alpha^i[a_\alpha]
    =
    \frac{C}{N}\sum_{\alpha\in i}a_\alpha
    -
    \theta_o\psi\,
    \frac{1}{N}\sum_{\alpha=1}^N a_\alpha,
    \label{eq:app_delta_E_definition}
\end{equation}
where $\alpha\in i$ means that the sum is restricted to training examples in class $i$.
Since $\mathrm{E}_\alpha[y_{\alpha i}]=1/C$, we have
\begin{equation}
    \mathrm{Cov}_\alpha^1(a_\alpha,y_{\alpha i})
    =
    \frac{1}{C}
    \Delta\mathrm{E}_\alpha^i[a_\alpha].
    \label{eq:app_cov1_deltaE_relation}
\end{equation}
The output therefore becomes
\begin{align}
f_i(x_*;\epsilon)
=&\
\frac{\theta_*\psi}{C}
+
\sqrt{\frac{N/C}{d}}
\frac{\sqrt{V_d}}{\theta_d-\theta_o}
\hat{\xi}_i
\notag \\
&+
\frac{\epsilon N/C}{\theta_d-\theta_o}
\Bigg(
    \left(
        S_{\chi *}^{1\rightarrow L}
        -
        \theta_*\psi S_{\chi\chi}^{1\rightarrow L}
    \right)
    \Delta\mathrm{E}_\alpha^i
    \left[
        \frac{\partial K_{\alpha\alpha}^{(1)}}{\partial\epsilon}
    \right]
\notag \\
&\qquad\qquad\qquad
    +
    T_{*\chi}^{1\rightarrow L}
    \Delta\mathrm{E}_\alpha^i
    \left[
        \frac{\partial K_{*\alpha}^{(1)}}{\partial\epsilon}
    \right]
\notag \\
&\qquad\qquad\qquad
    -
    \theta_*\psi
    \Bigg(
        T_{\chi\chi}^{1\rightarrow L}
        \mathrm{E}_\alpha
        \left[
            \Delta\mathrm{E}_\beta^i
            \left[
                \frac{\partial K_{\alpha\beta}^{(1)}}{\partial\epsilon}
            \right]
        \right]
\notag \\
&\qquad\qquad\qquad\qquad\quad
        +
        S_{\chi\chi}^{1\rightarrow L}
        (1-\theta_o\psi)
        \mathrm{E}_\alpha
        \left[
            \frac{\partial K_{\alpha\alpha}^{(1)}}{\partial\epsilon}
        \right]
    \Bigg)
\Bigg)
\notag \\
&+
\mathcal{O}(\epsilon N^0)
+
\mathcal{O}(\epsilon^2)
+
\mathcal{O}(N^0/\sqrt d)
+
\mathcal{O}(\epsilon N/\sqrt d)
+
\mathcal{O}(N/d).
\label{eq:app_general_classification_output}
\end{align}

We now impose the normalization used in the main text,
\begin{equation}
    \frac{1}{d}\sum_{j=1}^d x_{j\alpha}=0,
    \qquad
    \frac{1}{d}\sum_{j=1}^d x_{j\alpha}^2=1,
    \label{eq:app_data_normalization}
\end{equation}
for both training and test data.

Here we consider two general noise models. Additive noise with any mean-zero, unit-variance noise variable,
\begin{equation}
    \tilde{x}_{j\chi}
    =
    \epsilon x_{j\chi}
    +
    (1-\epsilon)\xi_{j\chi},
    \qquad
    \mathbb{E}[\xi_{j\chi}]=0,
    \qquad
    \mathbb{E}[\xi_{j\chi}^2]=1,
    \label{eq:app_additive_noise_model}
\end{equation}
for which the first-layer derivatives are
\begin{equation}
    \frac{\partial K_{*\chi}^{(1)}}{\partial\epsilon}
    =
    \frac{C_w}{d}
    \sum_{j=1}^d x_{j*}x_{j\chi},
    \quad
    \frac{\partial K_{\chi\chi'}^{(1)}}{\partial\epsilon}
    =
    -2C_w\delta_{\chi\chi'},
    \label{eq:app_additive_derivatives}
\end{equation}
and replacement noise with an arbitrary replacement distribution $u$,
\begin{equation}
    \tilde{x}_{j\chi}
    =
    (1-b_{j\chi})x_{j\chi}
    +
    b_{j\chi}u_{j\chi},
    \qquad
    \Pr(b_{j\chi}=0)=\epsilon,
    \label{eq:app_replacement_noise_model}
\end{equation}
for which the normalized data imply
\begin{align}
    \frac{\partial K_{*\chi}^{(1)}}{\partial\epsilon}
    &=
    \frac{C_w}{d}
    \sum_{j=1}^d x_{j*}x_{j\chi},
    \\
    \frac{\partial K_{\chi\chi'}^{(1)}}{\partial\epsilon}
    &=
    C_w\delta_{\chi\chi'}
    \left(
        1-\mathbb{E}[u^2]
    \right)
    -
    2C_w(1-\delta_{\chi\chi'})
    \mathbb{E}[u]^2.
    \label{eq:app_replacement_derivatives}
\end{align}
Thus, for both additive noise and replacement noise, the only class-dependent data dependence in Eq.~\eqref{eq:app_general_classification_output} comes from
\begin{equation}
    \Delta\mathrm{E}_\alpha^i
    \left[
        \frac{\partial K_{*\alpha}^{(1)}}{\partial\epsilon}
    \right]
    =
    C_w
    \Delta\mathrm{E}_\alpha^i
    \left[
        \frac{1}{d}
        \sum_{j=1}^d x_{j*}x_{j\alpha}
    \right].
    \label{eq:app_only_data_dependence}
\end{equation}
All other quantities in Eq.~\eqref{eq:app_general_classification_output}, including $\theta_*$, $\theta_d$, $\theta_o$, $S$, $T$, and $V_d$, are constants determined only by the architecture and the noise distribution.
This is because, under the normalization Eq.~\eqref{eq:app_data_normalization}, the pure-noise kernels depend on the data only through fixed quantities such as $\frac{1}{d}\sum_j x_{j\alpha}$ and $\frac{1}{d}\sum_j x_{j\alpha}^2$.

Therefore, for either noise model,
\begin{align}
f_i(x_*;\epsilon)
=&\
\mathrm{const}
+
A^{(L)}
\sqrt{\frac{N/C}{d}}\,
\hat{\xi}_i
+
B^{(L)}
\epsilon
\frac{N}{C}
\Delta\mathrm{E}_\alpha^i
\left[
    \frac{1}{d}
    \sum_{j=1}^d x_{j*}x_{j\alpha}
\right]
\notag \\
&+
\mathcal{O}(\epsilon N^0)
+
\mathcal{O}(\epsilon^2)
+
\mathcal{O}(N^0/\sqrt d)
+
\mathcal{O}(\epsilon N/\sqrt d)
+
\mathcal{O}(N/d),
\label{eq:app_general_centroid_form}
\end{align}
where $A^{(L)}$ and $B^{(L)}$ are architecture- and noise distribution-dependent constants.

Finally, to obtain the expression quoted in the main text, note that when $\theta_o\neq0$,
\begin{equation}
    \theta_o\psi
    =
    \frac{\theta_o N}{\theta_d-\theta_o+\theta_o N}
    =
    1+\mathcal{O}(N^{-1}).
    \label{eq:app_theta_o_psi_largeN}
\end{equation}
The exceptional case $\theta_o=0$ can occur in special symmetry-preserving settings, such as when the activation is chosen to be odd and $C_b=0$.
This is a non-generic degeneracy rather than a fundamental obstruction.
Indeed, adding any positive bias variance $C_b>0$, however small, breaks this exact odd symmetry at the level of the kernel recursion and generically produces a nonzero off-diagonal pure-noise kernel $\theta_o$.
Thus the case $\theta_o=0$ can be avoided by an arbitrarily small and standard modification of the initialization.
In the generic case $\theta_o\neq 0$, we therefore have
\begin{equation}
    \Delta\mathrm{E}_\alpha^i[a_\alpha]
    =
    \frac{C}{N}\sum_{\alpha\in i}a_\alpha
    -
    \frac{1}{N}\sum_{\alpha=1}^N a_\alpha
    +
    \mathcal{O}(N^{-1}).
    \label{eq:app_deltaE_largeN}
\end{equation}
Defining the empirical class mean (centroid) and global mean by
\begin{equation}
    \bar{x}_j^i
    =
    \frac{C}{N}\sum_{\alpha\in i}x_{j\alpha},
    \qquad
    \bar{x}_j
    =
    \frac{1}{N}\sum_{\alpha=1}^N x_{j\alpha},
    \label{eq:app_class_means}
\end{equation}
we obtain Eq.~\eqref{eq:maintxt_final_gaussian_output} of the main text, 
\begin{align}
f_i(x_*;\epsilon)
=&\
\mathrm{const}
+
A^{(L)}
\sqrt{\frac{N/C}{d}}\,
\hat{\xi}_i
+
B^{(L)}
\epsilon
\frac{N}{C}
\frac{1}{d}
\sum_{j=1}^d x_{j*}
\left(
    \bar{x}_j^i-\bar{x}_j
\right)
\notag \\
&+
\mathcal{O}(\epsilon N^0)
+
\mathcal{O}(\epsilon^2)
+
\mathcal{O}(N^0/\sqrt d)
+
\mathcal{O}(\epsilon N/\sqrt d)
+
\mathcal{O}(N/d).
\label{eq:app_final_high_noise_output}
\end{align}
Using $\epsilon=1-p$ gives the high-noise expression stated in the main text.
The leading class-dependent term is therefore the overlap between the clean test point and the centered empirical class centroid.
Consequently, after averaging over the finite-$d$ fluctuations, the network implements a nearest-class-mean rule in the high-noise limit.

\section{Nonnegativity of the coefficient multiplying the centroid}
\label{app:B_nonnegative}

For the centroid term in Eq.~\eqref{eq:maintxt_final_gaussian_output} to
have the correct orientation, its coefficient \(B^{(L)}\) must be
nonnegative. We show here that this is guaranteed by the normalization
convention and the unit-variance noise models adopted in the main text.

According to Eq.~\eqref{eq:app_general_classification_output}, the sign of
\(B^{(L)}\) is that of \(T^{(L)}_{*\chi}\), so it is enough to establish
\(T^{(L)}_{*\chi}\geq 0\). The recursion that generates
\(T^{(L)}_{*\chi}\) (Appendix~\ref{app:NTK}) involves only the
initialization constants \(C_b\geq 0\) and \(C_w>0\), the test--train
kernels \(K^{(\ell)}_{*\chi}\) and NTKs \(\Theta^{(\ell)}_{*\chi}\), and the
two-dimensional Gaussian expectations
\begin{equation}
    \langle h_* h_\chi\rangle_{K^{(\ell)}},
\end{equation}
in which \(h\) stands for the activation or one of its derivatives,
evaluated on a pair of jointly Gaussian pre-activations. The
nonnegativity of \(T^{(L)}_{*\chi}\) follows once each of these
ingredients is shown to be nonnegative at every layer.

The essential observation concerns the Gaussian expectation itself. Let
\((z_*,z_\chi)\) be jointly Gaussian with zero mean and covariance
\begin{equation}
    K
    =
    \begin{pmatrix}
    K_{**} & K_{*\chi} \\
    K_{*\chi} & K_{\chi\chi}
    \end{pmatrix},
\end{equation}
and suppose the two pre-activations share a common variance and a
nonnegative correlation,
\begin{equation}
    K_{**}=K_{\chi\chi}>0,
    \qquad
    K_{*\chi}\geq 0 .
    \label{eq:app_positivity_conditions}
\end{equation}
Because the two marginals have the common variance \(K_{**}\), we may
rescale to standard normal variables, writing \(z_*=\sqrt{K_{**}}\,X\) and
\(z_\chi=\sqrt{K_{**}}\,Y\) with \((X,Y)\) jointly standard normal. Their
correlation is inherited from the off-diagonal entry of \(K\),
\begin{equation}
    \rho=\frac{K_{*\chi}}{K_{**}}\in[0,1],
\end{equation}
where \(\rho\geq 0\) follows from \(K_{*\chi}\geq 0\), and \(\rho\leq 1\)
because \(K\) is a valid covariance matrix. We now expand the activation in
the Hermite polynomials \(H_n\), orthogonal with respect to the standard
Gaussian weight (\(H_0=1\), \(H_1=x\), \(H_2=x^2-1\), and so
on)~\cite{abramowitz_stegun,dlmf},
\begin{equation}
    h\!\left(\sqrt{K_{**}}\,x\right)
    =
    \sum_{n=0}^{\infty} c_n\,H_n(x).
\end{equation}
Using the bivariate identity
\(\E[H_m(X)H_n(Y)]=\delta_{mn}\,n!\,\rho^{\,n}\), the expectation collapses
to a single sum over the diagonal,
\begin{equation}
    \langle h_* h_\chi\rangle_{K}
    =
    \E\!\left[h(z_*)h(z_\chi)\right]
    =
    \sum_{n=0}^{\infty} c_n^2\,n!\,\rho^{\,n}
    \geq 0 ,
    \label{eq:app_hermite_positivity}
\end{equation}
every term of which is nonnegative because \(\rho\geq 0\). A common
variance together with a nonnegative correlation is therefore sufficient to
render \(\langle h_* h_\chi\rangle_{K}\) nonnegative for any activation \(h\) 
square-integrable with respect to a Gaussian measure; in particular this holds for
\(h=\sigma,\sigma',\sigma''\).

It remains to verify that the conditions
Eq.~\eqref{eq:app_positivity_conditions} are met at every layer. At the
first layer the kernel is fixed directly by the input overlaps,
\begin{equation}
    K^{(1)}
    =
    C_b
    +
    C_w
    \begin{pmatrix}
    \frac{1}{d}\sum_j x_{j*}^2
    &
    \E[u]\,\frac{1}{d}\sum_j x_{j*}
    \\[4pt]
    \E[u]\,\frac{1}{d}\sum_j x_{j*}
    &
    \E[u^2]
    \end{pmatrix},
\end{equation}
the constant \(C_b\) being added to every entry. The normalization
\begin{equation}
    \frac{1}{d}\sum_j x_{j*}=0,
    \qquad
    \frac{1}{d}\sum_j x_{j*}^2=1
\end{equation}
annihilates the off-diagonal signal and fixes the test-point variance to
unity, leaving \(K^{(1)}_{*\chi}=C_b\geq 0\). The diagonal entries,
\(C_b+C_w\) and \(C_b+C_w\,\E[u^2]\), coincide precisely when
\(\E[u^2]=1\) --- which is the case for both noise models employed here,
since replacement noise with \(u\sim\mathrm{Unif}[-\sqrt{3},\sqrt{3}]\) has
\(\E[u]=0\) and \(\E[u^2]=1\), and the additive Gaussian noise is taken
with zero mean and unit variance. The layer-one kernel thus satisfies
Eq.~\eqref{eq:app_positivity_conditions}, with \(K^{(1)}_{**}>0\) assured by
\(C_w>0\).

This structure is inherited under the layer-to-layer recursion. The
diagonal entries evolve according to
\(K^{(\ell+1)}_{\alpha\alpha}=C_b+C_w\langle\sigma^2\rangle_{K^{(\ell)}_{\alpha\alpha}}\),
which depends on the layer-\(\ell\) kernel only through the single diagonal
entry \(K^{(\ell)}_{\alpha\alpha}\); the two diagonal entries are therefore
governed by the same map, so equal diagonals at layer \(\ell\) yield equal
diagonals at layer \(\ell+1\). The off-diagonal entry evolves according to
\begin{equation}
    K^{(\ell+1)}_{*\chi}
    =
    C_b
    +
    C_w
    \langle\sigma_*\sigma_\chi\rangle_{K^{(\ell)}}
    \geq C_b \geq 0 ,
\end{equation}
where the inequality holds because the quantity
\(\langle\sigma_*\sigma_\chi\rangle_{K^{(\ell)}}\) is itself nonnegative:
this is precisely Eq.~\eqref{eq:app_hermite_positivity} with \(h=\sigma\),
applied to the layer-\(\ell\) kernel, which by hypothesis has equal
diagonals and a nonnegative off-diagonal entry. Both conditions
Eq.~\eqref{eq:app_positivity_conditions} are thus passed from layer
\(\ell\) to layer \(\ell+1\), and since they hold at the first layer they
hold at every layer up to \(L\). The same reasoning applied to the NTK
recursion
\(\Theta^{(\ell+1)}_{*\chi}=K^{(\ell+1)}_{*\chi}+C_w\langle\sigma'_*\sigma'_\chi\rangle_{K^{(\ell)}}\Theta^{(\ell)}_{*\chi}\),
beginning from \(\Theta^{(1)}_{*\chi}=K^{(1)}_{*\chi}=C_b\geq 0\), shows that
\(\Theta^{(\ell)}_{*\chi}\geq 0\) throughout.

With the common-variance and nonnegative-correlation properties secured at
every layer, the Gaussian expectations
\(\langle h_* h_\chi\rangle_{K^{(\ell)}}\) --- for
\(h=\sigma,\sigma',\sigma''\) alike --- are all nonnegative, as are
\(K^{(\ell)}_{*\chi}\) and \(\Theta^{(\ell)}_{*\chi}\). Every term of the
recursion for \(T^{(L)}_{*\chi}\) is then a sum of products of nonnegative
quantities, so that
\begin{equation}
    T^{(L)}_{*\chi}\geq 0,
    \qquad\text{and hence}\qquad
    B^{(L)}\geq 0 .
\end{equation}
The centroid term in Eq.~\eqref{eq:maintxt_final_gaussian_output} therefore
carries a nonnegative coefficient under our normalization convention and
unit-variance noise models.

\section{Empirical validation of the centroid model for additive-Gaussian noise}\label{app:additive-gaussian}

\begin{figure*}[h]
    \centering
    \includegraphics[width=1.0\linewidth]{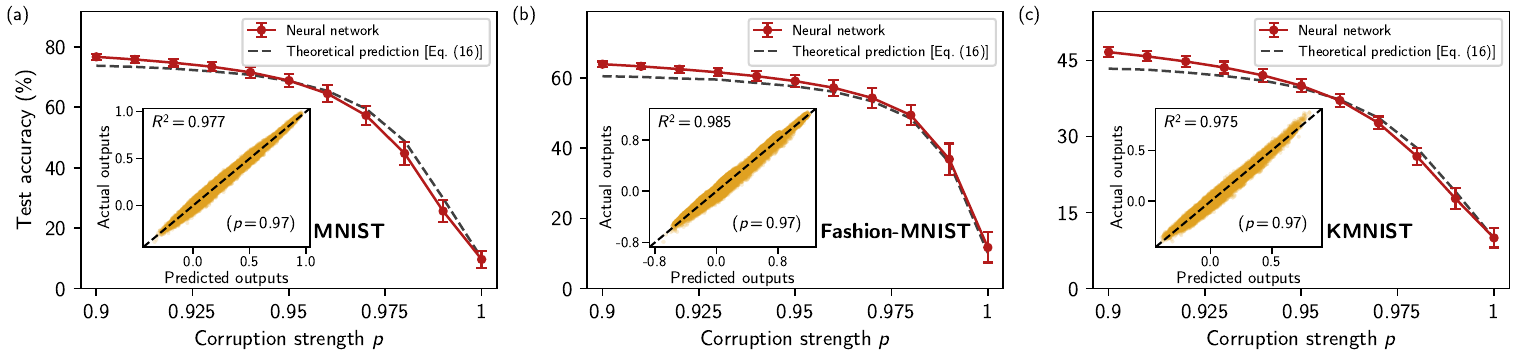}
    \caption{Numerical verification of the high-noise theory. All networks used to create the data in this figure are width-2048, 3-layer \(\mathrm{erf}\) MLPs, trained with MSE loss on noisy MNIST, FashionMNIST, and KMNIST with \(N=4000\). Main panel: mean clean-test accuracy versus additive-Gaussian-noise strength \(p\) for a nested ensemble of 20 noisy training sets and 10 MLPs per noisy set. Error bars show one standard deviation. The blue curve is the mean clean-test accuracy of the fitted effective model in Eq.~\eqref{eq:maintxt_output_fit}. Inset: output-level comparison at \(p=0.97\), using a nested ensemble of 100 noisy training sets and 50 MLPs per noisy set. Each point corresponds to one clean test image and one candidate class, comparing the ensemble-mean network output with the fitted prediction from Eq.~\eqref{eq:maintxt_output_fit}. The black dashed line in the inset indicates perfect agreement.}
    \label{fig:actual_vs_predicted_additive_gaussian}
\end{figure*}

We now show that the same high-noise picture extends to additive-Gaussian corruption. In the insets of Fig.~\ref{fig:actual_vs_predicted_additive_gaussian}, we compare the ensemble-averaged MLP logits on the full MNIST, FashionMNIST, and KMNIST test sets with the prediction obtained from the mean of Eq.~\eqref{eq:maintxt_output_fit} for noise strength $p=0.97$. Using a global fit for the coefficients \(a\) and \(b\), we observe excellent agreement. This indicates that, once averaged over noise realizations, the network outputs are well described by the mean centroid-based predictor in the additive-Gaussian setting as well. The main panels of Fig.~\ref{fig:actual_vs_predicted_additive_gaussian} compare the clean-test accuracy of the empirical ensemble against that of the full effective model in Eq.~\eqref{eq:maintxt_output_fit}, including the fluctuation term set by the fitted coefficient \(c\). The close match shows that the centroid description accounts not only for the average logits, but also for the noise-induced variations that determine classification performance. In particular, it reproduces the decline in clean-test accuracy as \(p\) increases. The figure also shows where the empirical networks begin to deviate from the leading-order centroid prediction, indicating the growing importance of higher-order terms in \((1-p)\). In our experiments, these deviations become visible for \(p \lesssim 0.9\).

\begin{figure*}[th]
    \centering
    \includegraphics[width=\linewidth]{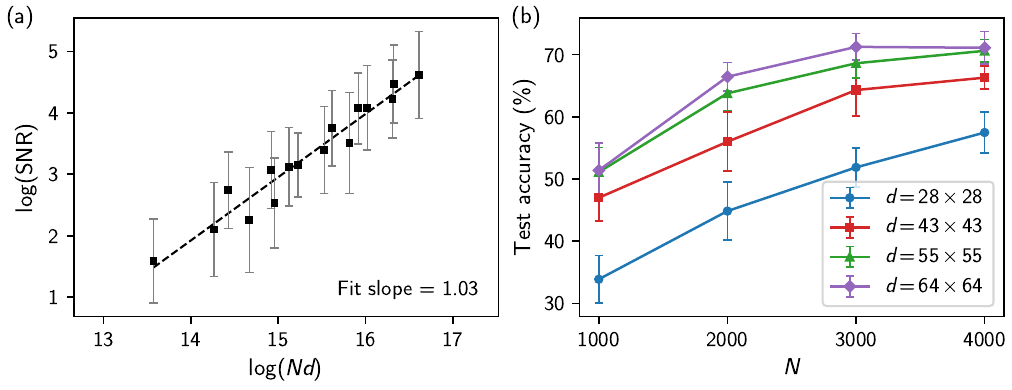}
    \caption{
    Empirical $N,d$ scaling at fixed additive-Gaussian noise probability $p=0.97$ and $C=10$ classes for width-2048, 3-layer $\mathrm{erf}$ MLPs trained with MSE loss on MNIST.
    (a)~Mean $\log(\mathrm{SNR})$ versus $\log(Nd)$ over 500 randomly sampled clean-test examples. Per-example SNR is computed from true-class logits averaged over 10 initializations per noisy training dataset, using the squared mean divided by the sample variance across 20 noisy training datasets. Squares and error bars show the mean and one standard deviation of log(SNR) across test examples; the dashed line is a linear fit to the means.
    (b)~Mean clean-test accuracy versus $N$ at fixed $d$, using predictions from initialization-averaged logits. Means and one-standard-deviation error bars are computed across the 20 noisy training datasets. Larger feature dimensions $d$ are obtained by resizing MNIST images using bilinear interpolation.}
    \label{fig:test_acc_vs_N_d_additive_gaussian}
\end{figure*}

We next examine the $N,d$ dependence of the empirical SNR and test accuracy under additive-Gaussian noise, using MNIST at fixed $p=0.97$ and $C=10$. Figure~\figref{fig:test_acc_vs_N_d_additive_gaussian}{a} compares the mean log SNR with $\log(Nd)$, using the same estimator and averaging procedure as in the main text. This provides a direct test of the $\mathrm{SNR}\propto Nd$ dependence suggested by Eq.~\eqref{eq:maintxt_snr_scaling}. Panel (b) shows that clean-test accuracy increases systematically with both $N$ and $d$, consistent with the proposed scaling.

\section{Test accuracy over the full noise range}
\label{app:test_acc_p=0-1}

\begin{figure*}[th]
    \centering
    \includegraphics[width=\linewidth]{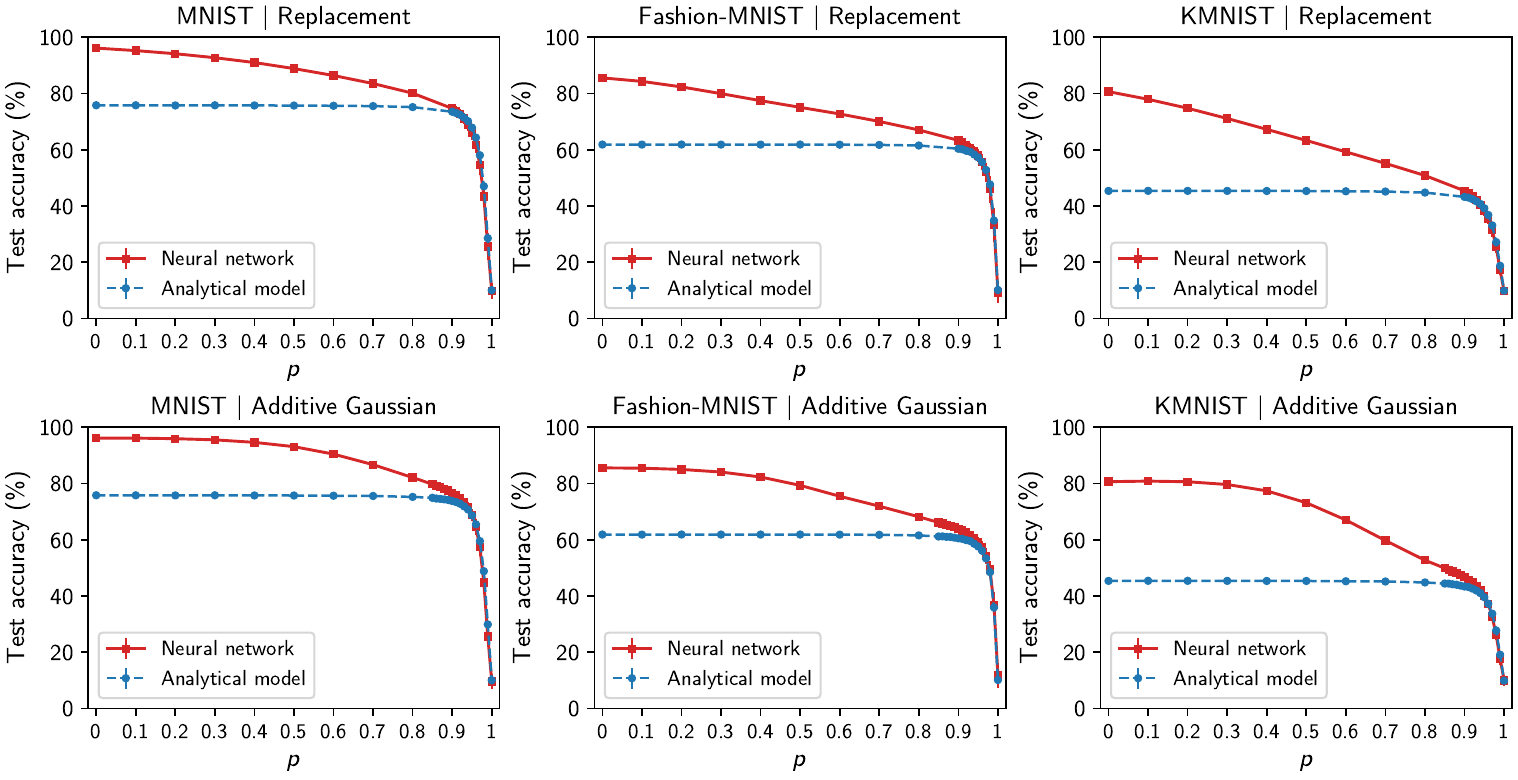}
    \caption{Clean-test accuracy over the full noise range $p\in[0,1]$ for replacement noise (top row) and additive-Gaussian noise (bottom row). Columns show MNIST, Fashion-MNIST, and KMNIST, from left to right. Red solid curves show empirical neural-network accuracy; blue dashed curves show the centroid-based effective model of Eq.~\eqref{eq:maintxt_output_fit}, calibrated at $p=0.97$ and evaluated across the full range. Networks are width-2048, 3-layer $\mathrm{erf}$ MLPs trained with MSE loss, $N=4000$, and $d=28\times28$. Ensemble averaging and all other hyperparameters match the accuracy panels of Figs.~\ref{fig:actual_vs_predicted} and~\ref{fig:actual_vs_predicted_additive_gaussian} for the respective noise types. Error bars show one standard deviation.}
    \label{fig:centroid_vs_actual_depth=3_p=0-1}
\end{figure*}

Figure~\ref{fig:centroid_vs_actual_depth=3_p=0-1} extends the centroid-model comparison to the full range $p\in[0,1]$. The replacement-noise and additive-Gaussian experiments use the same hyperparameters and evaluation procedures as Figs.~\ref{fig:actual_vs_predicted} and~\ref{fig:actual_vs_predicted_additive_gaussian}, respectively. The extended curves show how the agreement in the high-noise regime gives way to a growing accuracy advantage for the neural networks as $p$ decreases. This comparison illustrates the limits of the leading-order centroid description and the additional classification performance attained by the networks outside that regime.

\section{Validation for a deeper ReLU architecture}
\label{app:architectural_universality}

\begin{figure*}[th]
    \centering
    \includegraphics[width=\linewidth]{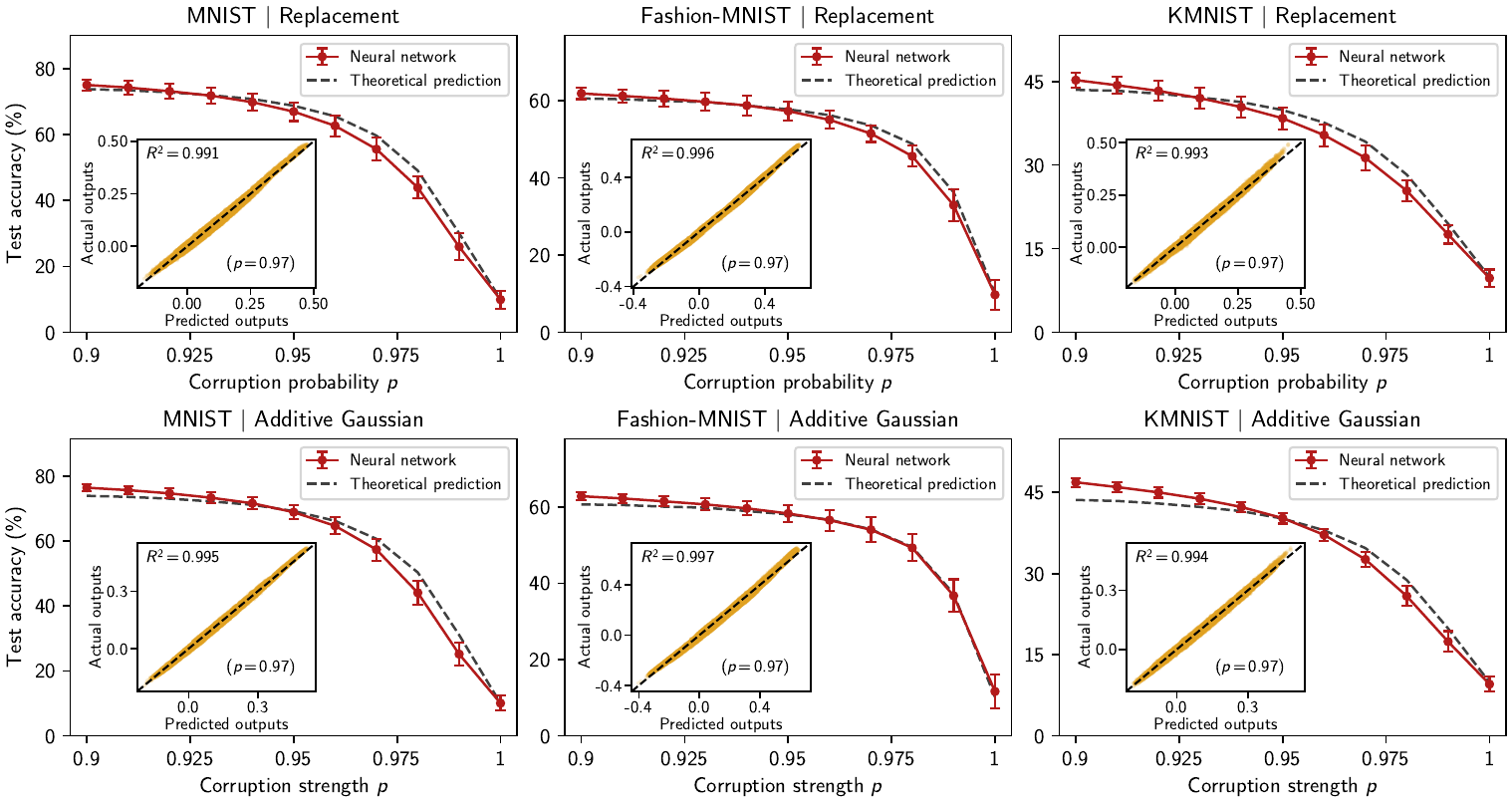}
    \caption{Comparison with depth-$5$, width-2048 ReLU MLPs trained with MSE loss under replacement noise (top row) and additive-Gaussian noise (bottom row). Columns show MNIST, Fashion-MNIST, and KMNIST, from left to right. Main panels: mean clean-test accuracy versus $p$ for the neural networks (red solid curves) and the fitted centroid-based effective model of Eq.~\eqref{eq:maintxt_output_fit} (blue dashed curves). Error bars show one standard deviation. Insets: ensemble-mean output logits versus fitted mean predictions at $p=0.97$, with each point representing a clean test example and candidate class; dashed diagonal lines indicate perfect agreement. Training uses $N=4000$ and $d=28\times28$. All settings other than depth and activation, including ensemble sizes and fitting procedures, match Figs.~\ref{fig:actual_vs_predicted}~and ~\ref{fig:actual_vs_predicted_additive_gaussian} and for the respective noise types.}
    \label{fig:centroid_vs_actual_depth=5}
\end{figure*}

We repeat the output-level and test-accuracy comparisons we performed in Figs.~\ref{fig:actual_vs_predicted} and~\ref{fig:actual_vs_predicted_additive_gaussian} for depth-$5$ ReLU MLPs. All other hyperparameters and evaluation procedures remain unchanged for the respective noise types, and the effective-model coefficients are fitted using the same procedure. As shown in Fig.~\ref{fig:centroid_vs_actual_depth=5}, the centroid-based model continues to describe the ensemble-mean logits and broadly captures the high-noise accuracy curves. These results provide an additional empirical check that the centroid description extends beyond the architecture used in the main text.

\section{Dependence of test accuracy on activation}\label{app:test_acc_activation}

In this appendix we discuss how the high-noise prediction accuracy depends on the activation function.
For analytical simplicity, we specialize the general result of Appendix~\ref{app:NTK} to additive-Gaussian noise.
This choice makes the fluctuation covariance particularly transparent, while still capturing the qualitative mechanism expected for general i.i.d.\ corruption models: the class-dependent signal is controlled by the centroid overlap derived in Eq.~\eqref{eq:app_final_high_noise_output}, whereas the activation function enters only through architecture-dependent constants.

We use the same data normalization convention as in Appendix~\ref{app:NTK},
\begin{equation}
    \frac{1}{d}\sum_{j=1}^d x_{j\alpha}=0,
    \qquad
    \frac{1}{d}\sum_{j=1}^d x_{j\alpha}^2=1.
\end{equation}
Under this normalization, the pure-noise training--training and test-training kernels coincide at the level relevant for this calculation.
Thus
\begin{equation}
    S_{\chi *}^{1\rightarrow L}=S_{\chi\chi}^{1\rightarrow L}=S,
    \qquad
    T_{*\chi}^{1\rightarrow L}=T_{\chi\chi}^{1\rightarrow L}=T,
    \qquad
    \theta_*=\theta_o.
\end{equation}

For additive Gaussian noise, substituting the corresponding first-layer derivatives into the general high-noise output formula, Eq.~\eqref{eq:app_general_classification_output}, gives
\begin{align}
    f_i(x_*;\epsilon)
    =&\
    \frac{\theta_o\psi}{C}
    +
    \sqrt{\frac{N/C}{d}}
    \frac{\sqrt{V_d}}{\theta_d-\theta_o}
    \hat{\xi}_i
    \notag \\
    &+
    C_w
    \frac{\epsilon N/C}{\theta_d-\theta_o}
    \Bigg(
        -2S(1-\theta_o\psi)(1-2\theta_o\psi)
        +
        T\Delta\mathrm{E}^i_\alpha
        \left[
            \frac{1}{d}\sum_{j=1}^d x_{j*}x_{j\alpha}
        \right]
    \Bigg)
    \notag \\
    &+
    \mathcal{O}(\epsilon N^0)
    +
    \mathcal{O}(\epsilon^2)
    +
    \mathcal{O}(N^0/\sqrt d)
    +
    \mathcal{O}(\epsilon N/\sqrt d)
    +
    \mathcal{O}(N/d),
    \label{eq:final_gaussian_output_activation}
\end{align}
where
\begin{align}
    V_d
    =
    C_w^2
    \Bigg[
        &2S^2(1-\theta_o\psi)^2
        \left(
            1-\frac{4}{C}\theta_o\psi(1-\theta_o\psi)
        \right)
        \notag \\
        &+
        T^2
        \left(
            1-\frac{1}{C}\theta_o\psi(2-\theta_o\psi)
        \right)
    \Bigg].
    \label{eq:activation_V_d}
\end{align}
Here $\hat{\xi}_i$ denotes the Gaussian fluctuation described in Appendix~\ref{app:NTK}, and $\Delta\mathrm{E}^i_\alpha[\cdot]$ is the class-centered average defined in Eq.~\eqref{eq:app_delta_E_definition}.
The only term in Eq.~\eqref{eq:final_gaussian_output_activation} with classifying power is the centroid-overlap term
\begin{equation}
    \Delta\mathrm{E}^i_\alpha
    \left[
        \frac{1}{d}\sum_{j=1}^d x_{j*}x_{j\alpha}
    \right].
\end{equation}
All remaining terms are either class-independent shifts or random fluctuations.
Under the normalization and unit-variance noise conventions used here, the coefficient multiplying this centroid term has the correct nonnegative orientation; equivalently, \(T\geq 0\), as shown in Appendix~\ref{app:B_nonnegative}. Thus different activations change the magnitude of the signal and the fluctuation scale, but they do not reverse the orientation of the leading centroid decision rule in the setting considered here. 

For fixed data and noise level, activation-dependent robustness is therefore governed by the noise-to-signal ratio between the fluctuation scale and the coefficient of this centroid term. Equivalently, maximizing prediction accuracy corresponds to minimizing
\begin{equation}
    \frac{V_d}{T^2}.
    \label{eq:activation_snr_objective}
\end{equation}
Since the optimal activation should not depend sensitively on the number of classes, it is useful to examine the large-$C$ limit.
In this limit,
\begin{align}
    V_d
    =
    C_w^2
    \left[
        2S^2(1-\theta_o\psi)^2
        +
        T^2
    \right].
    \label{eq:activation_V_d_large_C}
\end{align}
Thus, at large $C$, the excess fluctuation beyond the unavoidable $T$ contribution is controlled by
\begin{equation}
    S^2(1-\theta_o\psi)^2.
\end{equation}
Consequently, for the generic case in which the pure-noise off-diagonal NTK is nonzero, the large-$N$ relation $\theta_o\psi=1+\mathcal{O}(N^{-1})$ implies that this term is suppressed by $\mathcal{O}(N^{-2})$. Thus, a broad class of activations should have nearly identical signal-to-noise ratios, and therefore similar test accuracy.
In other words, the leading high-noise classifier is not strongly controlled by the detailed choice of activation; the activation mainly changes subleading architecture-dependent prefactors multiplying the same centroid rule.

\begin{figure}[h]
    \centering
    \includegraphics[width=0.5\linewidth]{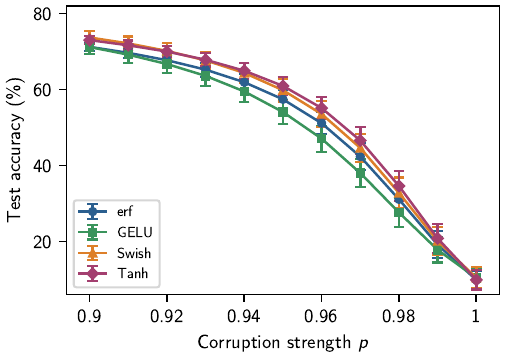}
    \caption{Mean test accuracy, with standard deviations shown as error bars, for an ensemble of 50 independent 3-layer, width 2048 neural networks with four different activations: erf, GELU, Swish, and Tanh. The networks are trained with MSE loss on a noisy subset of 4,000 MNIST data points corrupted with additive-Gaussian noise.}
    \label{fig:test_acc_activation}
\end{figure}

Figure~\ref{fig:test_acc_activation} tests this prediction numerically.
We compare four activations: erf, GELU, Swish, and Tanh.
The resulting test-accuracy curves are very similar across the full range of noise probabilities $p$.
Aside from a modest advantage for Tanh and Swish at the highest noise levels, all four activations attain comparable performance and display nearly the same dependence on $p$.


\section{Explanation of non-monotonic test accuracy variance}\label{app:test_acc_var}

Here we show that the non-monotonic dependence of the test-accuracy variance on the noise probability \(p\) follows directly from the general form of the output logits in Eq.~\eqref{eq:maintxt_final_gaussian_output}. For clarity, we consider the following toy model. The logit of the correct class is
\begin{equation}
    l_* = \hat{\xi}_* + a(1-p),
    \label{eq:toy_model_logit}
\end{equation}
where \(\hat{\xi}_* \sim \mathcal{N}(0,1)\) and \(a>0\) is a constant. The logits of the \(C-1\) incorrect classes are
\begin{equation}
    l_i = \hat{\xi}_i, \qquad i=1,\dots,C-1,
\end{equation}
where each \(\hat{\xi}_i \sim \mathcal{N}(0,1)\). We assume all Gaussian variables are independent. This captures the essential structure of Eq.~\eqref{eq:maintxt_final_gaussian_output}: the correct class receives a positive deterministic offset, reflecting its larger centroid overlap, while the incorrect classes do not. For simplicity, we take this offset to be the constant \(a(1-p)\), and we normalize the Gaussian fluctuations to zero mean and unit variance.

For a test input \(x_*\), define the correctness indicator
\begin{equation}
    \mathrm{Acc}(x_*) = 1\!\left\{ l_*(x_*) > \max_{1 \le i \le C-1} l_i(x_*) \right\}.
\end{equation}
The population mean test accuracy is therefore
\begin{equation}
    \mathbb{E}[\mathrm{Acc}]
    = \mathbb{E}_{x_*}\!\left[
    1\!\left\{ l_*(x_*) > \max_{1 \le i \le C-1} l_i(x_*) \right\}
    \right].
\end{equation}
Since \(\mathrm{Acc}(x_*)\) is an indicator variable, \(\mathrm{Acc}(x_*)^2=\mathrm{Acc}(x_*)\), and thus its population variance is
\begin{equation}
    \Var[\mathrm{Acc}]
    = \mathbb{E}[\mathrm{Acc}] - \mathbb{E}[\mathrm{Acc}]^2
    = \mathbb{E}[\mathrm{Acc}]\,\bigl(1-\mathbb{E}[\mathrm{Acc}]\bigr).
\end{equation}
Hence the variance is maximized when \(\mathbb{E}[\mathrm{Acc}] = 1/2\), so it is necessarily a non-monotonic function of the mean accuracy.

It therefore remains only to show that the mean accuracy is monotonic in \(p\). Conditioning on the correct-class noise \(\hat{\xi}_*=u\), the probability that the correct logit exceeds all \(C-1\) incorrect logits is
\begin{equation}
    \Phi\bigl(u+a(1-p)\bigr)^{C-1}.
\end{equation}
Averaging over \(u\sim\mathcal{N}(0,1)\) gives
\begin{equation}
    \mathbb{E}[\mathrm{Acc}]
    = \int_{-\infty}^{\infty} du \, \phi(u)\,
    \Phi\bigl(u+a(1-p)\bigr)^{C-1},
\end{equation}
where \(\phi\) and \(\Phi\) are the standard normal probability density function (PDF) and cumulative distribution function (CDF), respectively. Because \(\Phi\) is strictly increasing, this expression is monotonically decreasing in \(p\). It follows that \(\Var[\mathrm{Acc}]\), being of the form \(m(1-m)\) with \(m=\mathbb{E}[\mathrm{Acc}]\), is generically non-monotonic in \(p\). This proves the claim.

\begin{figure}[h]
    \centering
    \includegraphics[width=0.5\linewidth]{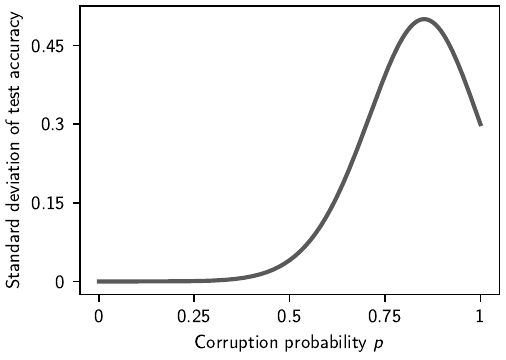}
    \caption{Standard deviation of the test accuracy of the toy model versus the noise probability $p$. We set the constants $a=10$ and $C=10$.}
    \label{fig:toy_model_stdv_test_acc}
\end{figure}

Figure~\ref{fig:toy_model_stdv_test_acc} shows the predicted standard deviation of the test accuracy in this toy model as a function of \(p\) for \(a=10\) and $C=10$. The non-monotonic behavior is clearly reproduced, showing that it arises directly from the competition between a decreasing correct-class offset and Gaussian logit fluctuations.

\end{document}